%% file: main.tex
\documentclass{vgtc}                          

\graphicspath{{figures/}{pictures/}{images/}{./}} 

\usepackage{times}                     

\onlineid{0}

\vgtccategory{Research}

\vgtcinsertpkg

\usepackage{wrapfig}
\usepackage{titlesec}
\usepackage{booktabs}
\usepackage{microtype}
\usepackage{listings}
\usepackage{gensymb}
\usepackage{algorithm}
\usepackage[noend]{algpseudocode}
\usepackage{subcaption}
\usepackage{amsmath}
\usepackage{dsfont}
\newcommand{\mathbb}[1]{\mathds{#1}}
\usepackage{tikz}
\usetikzlibrary{
    arrows.meta,
    bending,
    positioning,
    backgrounds
} 
\usepackage{graphicx}
\usepackage{dblfloatfix}
\usepackage{adjustbox}
\usepackage{balance}

\usepackage{mathptmx}            
\usepackage{wrapfig}    

\usepackage{setspace}

\newcommand{\mat}[1]{\ensuremath{\mathbf{#1}}}

\DeclareMathOperator*{\argmax}{argmax}
\newcommand{\argmaxF}{\mathop{\mathrm{argmax}}\limits}   

\definecolor{mplBlue}{HTML}{1f77b4}
\definecolor{mplOrange}{HTML}{ff7f0e}
\definecolor{mplGreen}{HTML}{2ca02c}
\definecolor{mplRed}{HTML}{d62728}
\definecolor{mplPurple}{HTML}{9467bd}
\definecolor{mplBrown}{HTML}{8c564b}
\definecolor{mplPink}{HTML}{e377c2}
\definecolor{mplGray}{HTML}{7f7f7f}
\definecolor{mplOlive}{HTML}{bcbd22}
\definecolor{mplCyan}{HTML}{17becf}

\definecolor{violet}{HTML}{8c1267}

\title{Feature Clock: High-Dimensional Effects in Two-Dimensional Plots}

\author{Olga Ovcharenko\thanks{e-mail: oovcharenko@ethz.ch}\\ %
        \scriptsize ETH Z{\"u}rich %
\and Rita Sevastjanova\thanks{e-mail: \{rita.sevastjanova$|$valentina.boeva\}@inf.ethz.ch}\\ %
     \scriptsize ETH Z{\"u}rich %
\and Valentina Boeva\footnotemark[2]\\ %
     \scriptsize ETH Z{\"u}rich %
}

\teaser{

    \input{figures/teaser}
}

\abstract{
    Humans struggle to perceive and interpret high-dimensional data. 
    Therefore, high-dimensional data are often projected into two dimensions for visualization.
    Many applications benefit from complex nonlinear dimensionality reduction techniques, but the effects of individual high-dimensional features are hard to explain in the two-dimensional space.
    Most visualization solutions use multiple two-dimensional plots, each showing the effect of one high-dimensional feature in two dimensions; this approach creates a need for a visual inspection of $k$ plots for a $k$-dimensional input space.
    Our solution, Feature Clock, provides a novel approach that eliminates the need to inspect these $k$ plots to grasp the influence of original features on the data structure depicted in two dimensions. 
    Feature Clock enhances the explainability and compactness of visualizations of embedded data and is available in an open-source Python library\footnote{\url{https://pypi.org/project/feature-clock}}. 
} 

\keywords{High-dimensional data, nonlinear dimensionality reduction, feature importance, visualization.}

\begin{document}

\firstsection{Introduction}

\maketitle
\input{1-Introduction}

\input{2-background}

\input{3-methods}

\input{4-evaluation}

\input{5-discussion}

\input{6-conclusions}

\input{7-acknowledgements}

\bibliographystyle{abbrv-doi}

\bibliography{bib}

\newpage

\input{8-appendix}

\end{document}

%% file: figures/teaser.tex
\tikzset{HDnodeBase/.style={draw=gray!50!white,fill=red, minimum height=6, minimum width=5, inner sep=0.0}}
\tikzset{HDnodeBlue/.style={HDnodeBase,fill=mplBlue}}
\tikzset{HDnodeRed/.style={HDnodeBase,fill=mplRed}}
\tikzset{HDnodePink/.style={HDnodeBase,fill=mplPink}}
\tikzset{HDnodeGreen/.style={HDnodeBase,fill=mplGreen}}
\tikzset{HDnodeCyan/.style={HDnodeBase,fill=mplCyan}}
\tikzset{HDnodeOrange/.style={HDnodeBase,fill=mplOrange}}
\tikzset{HDnodeOlive/.style={HDnodeBase,fill=mplOlive}}
\tikzset{HDnodeBrown/.style={HDnodeBase,fill=mplBrown}}
\tikzset{HDnodeGray/.style={HDnodeBase,fill=mplGray}}
\tikzset{HDnodeGray/.style={HDnodeBase,fill=mplGray}}
\tikzset{HDnodePurple/.style={HDnodeBase,fill=mplPurple}}

\tikzset{roundnode/.style={circle, draw=mplPink!60, fill=mplPink!5, very thick, minimum size=4mm, inner sep=0.0, font=\rmfamily,yshift=0.5cm}}

\newcommand{\nodeLabel}[1]{
    \begin{tikzpicture}[]
        \node[roundnode, minimum size=2mm]{\fontsize{5}{5}\selectfont{#1}};
    \end{tikzpicture}
}

\centering
\vspace{-0.5cm}
\begin{tikzpicture}[]
    \begin{scope}[yshift=0.4cm]
        \begin{scope}[xshift=-1.2cm] 
        \node[roundnode, xshift=-1.1cm]{a};
        \node[font=\rmfamily,yshift=0.8cm,anchor=south]{\fontsize{8}{8}\selectfont{\begin{tabular}{c}High dim. data \textbf{X}\end{tabular}}};

        \foreach \i in {1,...,21}
        {
            \pgfmathtruncatemacro{\y}{(\i - 1) / 7};
            \pgfmathtruncatemacro{\x}{(\i - 7 * \y)};
            \ifthenelse{\x = 1}{\node[HDnodeBlue]   at (0.21 * \x-0.84, 0.25 * \y + 0.25) {};}{}
            \ifthenelse{\x = 2}{\node[HDnodeRed]    at (0.21 * \x-0.84, 0.25 * \y + 0.25) {};}{}
            \ifthenelse{\x = 3}{\node[HDnodeGreen]  at (0.21 * \x-0.84, 0.25 * \y + 0.25) {};}{}
            \ifthenelse{\x = 4}{\node[HDnodePurple] at (0.21 * \x-0.84, 0.25 * \y + 0.25) {};}{}
            \ifthenelse{\x = 5}{\node[HDnodeGray]   at (0.21 * \x-0.84, 0.25 * \y + 0.25) {};}{}
            \ifthenelse{\x = 6}{\node[HDnodePink]   at (0.21 * \x-0.84, 0.25 * \y + 0.25) {};}{}
            \ifthenelse{\x = 7}{\node[HDnodeOrange] at (0.21 * \x-0.84, 0.25 * \y + 0.25) {};}{}
        }
    \end{scope}

    \begin{scope}[yshift=-0.9cm,xshift=-1.2cm] 
        \node[font=\rmfamily,yshift=0.5cm, anchor=south]{\fontsize{8}{8}\selectfont{\begin{tabular}{c} Low dim. data \textbf{Y} \end{tabular}}};

        \foreach \i in {1,...,6}
        {
            \pgfmathtruncatemacro{\y}{(\i - 1) / 2};
            \pgfmathtruncatemacro{\x}{(\i - 2 * \y)};
            \ifthenelse{\x = 1}{\node[HDnodeBrown]   at (0.21 * \x-0.21, 0.25 * \y) {};}{}
            \ifthenelse{\x = 2}{\node[HDnodeCyan]    at (0.21 * \x-0.21, 0.25 * \y) {};}{}
        }
    \end{scope}    
    \end{scope}

    \begin{scope}[xshift=1.5cm] 
        \node[roundnode, xshift=-0.65cm, yshift=0.8cm]{b};
        \node[font=\rmfamily,yshift=0.8cm, xshift=0.9cm, anchor=south]{\fontsize{8}{8}\selectfont{\begin{tabular}{c} Define optimization \\ problem \end{tabular}}};
        \node[font=\rmfamily,anchor=south,xshift=0.1cm,yshift=-0.4cm]{\fontsize{8}{8}\selectfont{
        \setlength\tabcolsep{0.pt}%
        \begin{tabular}{rl} 
        {\color{violet}$\mat{y_\theta}$} &$= \mat{X}\beta_\theta$ \\[0.17cm]
        $\theta^j$ &$= \argmaxF_{\theta \in [0..180\degree)} \lvert\beta^j\rvert$ \\[-0.07cm]
        \end{tabular}
        }};

         \begin{scope}[scale=1.6, xshift=.9cm, yshift=-0.1cm]
            \draw [gray,-{Latex[length=1.5mm]}, line width=0.2mm] (-0.1,0) -- (0.85,0);
            \draw[gray,-{Latex[length=1.5mm]}, line width=0.2mm] (0,-0.1) -- (0,0.6);

    
            \draw[line width=0.3mm, dashed] (-0.07,-0.04) -- (0.77,0.44);
            
            \draw[thick, ]
            (0.2296,0) .. controls (0.2277,0.0519)..
            (0.1890,0.0945);
            \node[anchor=west] at(0.3277,0.119){\fontsize{8}{8}\selectfont{$\theta$}};
            \draw[gray] (0.15,0.55)            -- (0.35,0.2);
            \draw[gray] (0.7+0.044,0.4-0.077)  -- (0.7,0.4);
            \draw[gray] (0.7-0.21-0.088,0.4-0.12+0.14) -- (0.7-0.21,0.4-0.12);
        
            \filldraw[violet] (0.35,0.2) circle (1pt);
            \filldraw[violet] (0.7,0.4) circle (1pt);
            \filldraw[violet] (0.7-0.21,0.4-0.12) circle (1pt);

            \filldraw[gray] (0.15,0.55) circle (1pt);
            \filldraw[gray] (0.7+0.066,0.4-0.0955) circle (1pt);
            \filldraw[gray] (0.7-0.21-0.088,0.4-0.12+0.14) circle (1pt);
            
        \end{scope}
    \end{scope}

    \begin{scope}[xshift=1.5cm] 
    \end{scope}

    \begin{scope}[xshift=10cm] 
        \node[roundnode, xshift=-1.0cm, yshift=0.8cm]{d};
        \node[font=\rmfamily,yshift=0.8cm, xshift=0.55cm, anchor=south]{\fontsize{8}{8}\selectfont{\begin{tabular}{c} Filter coefficients  {\color{mplGreen}$\beta_{\theta}$}\\ with p-values $\ge0.05$\end{tabular}}};
        \node[xshift=0.35cm, yshift=0.2cm]{\includegraphics[scale=0.07]{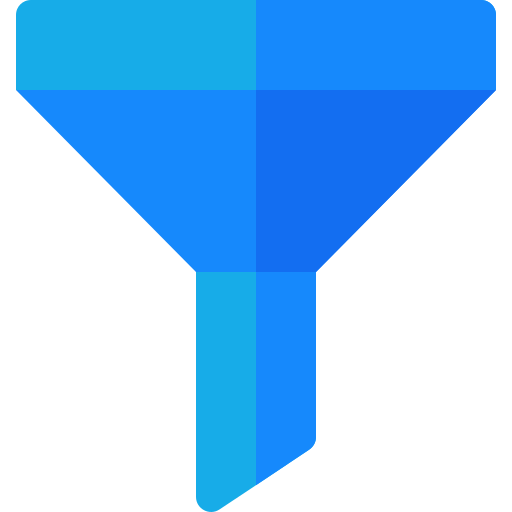}};
    \end{scope}

    \begin{scope}[xshift=5.8cm] 
        \node[roundnode, xshift=-1.05cm, yshift=0.8cm]{c};
        \node[font=\rmfamily,yshift=0.8cm, xshift=0.55cm, anchor=south]{\fontsize{8}{8}\selectfont{\begin{tabular}{c} Find largest coefficient \\for each feature $j$\end{tabular}}};

        \begin{scope}[xshift=0.3cm]
        \node[font=\rmfamily,yshift=-0.35cm, xshift=-0.0cm, anchor=south]{
        \setlength\tabcolsep{0.pt}%
        \fontsize{8}{8}\selectfont{\begin{tabular}{rl} {
        $($\color{mplGreen}$\beta_{\theta^j}^{j}$}$)^2$&$=(${\color{mplBrown}$\beta_{0\degree}^{j}$}$)^2+(${\color{mplCyan}$\beta_{90\degree}^{j}$}$)^2$  \vspace{0.14cm} \\
        $\theta^j$&$= \arctan(${\color{mplCyan}$\beta_{90\degree}^{j}$}$ / ${\color{mplBrown}$\beta_{0\degree}^{j}$}$)$  
        \end{tabular}}};

        \begin{scope}[scale=1.6, xshift=1.2cm, yshift=-0.1cm]
            \draw [gray,-{Latex[length=1.5mm]}, line width=0.2mm] (-0.1,0) -- (0.9,0);
            \draw[gray,-{Latex[length=1.5mm]}, line width=0.2mm] (0,-0.1) -- (0,0.6);

            \draw [gray,dotted, line width=0.16mm] (0,0.4) -- (0.7,0.4);
            \draw[gray,dotted,line width=0.16mm] (0.7,0) -- (0.7,0.4);
    
            \draw[mplCyan,-{Latex[length=2mm]}, line width=0.5mm] (0,0) -- (0,0.4);
            \draw[mplBrown, -{Latex[length=2mm]}, line width=0.5mm] (0,0) -- (0.7,0.0);
            \draw[mplGreen, -{Latex[length=2mm]}, line width=0.5mm] (0,0) -- (0.7,0.4);

            \draw[thick, ]
            (0.2296,0) .. controls (0.2277,0.0519)..
            (0.1890,0.0945);
            \node[anchor=west] at(0.3277,0.119){\fontsize{8}{8}\selectfont{$\theta^j$}};

            \filldraw[gray, fill=none] (0.35,0.2) circle (0.4);
            
        \end{scope}
        \end{scope}
    \end{scope}

    \begin{scope}[xshift=13.0cm, yshift=0.3cm] 
        \node[xshift=0.35cm]{\includegraphics[scale=0.8]{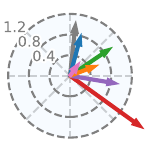}};
        \node[font=\rmfamily,yshift=0.8cm, xshift=0.35cm, anchor=south]{\fontsize{8}{8}\selectfont{\begin{tabular}{c} \textbf{Feature Clock}\end{tabular}}};
        \node[roundnode, xshift=-0.7cm,yshift=-1.1cm]{e};
    \end{scope}

    \begin{scope}[on background layer] 
        \node[fill=white!90!mplCyan, minimum height=2.3cm, minimum width=11.6cm,xshift=6.2cm, yshift=0.5cm,rounded corners=0.5cm]{};
    \end{scope}

\end{tikzpicture}

\vspace{-0.30cm}
\caption[teaser]{Feature Clock uses {\nodeLabel{a}} high \textbf{X} and low \textbf{Y} dimensional data.~\nodeLabel{b} 
We define the feature contribution as a coefficient of the linear regression (LR) between \textbf{X} and $\mathbf{y_\theta}$ (\textbf{Y} projected at angle $\theta$). The optimization goal is to find the angle $\theta^j$ at which the coefficient $\beta_{\theta^j}^j$ of feature $j$ is maximized. \nodeLabel{c} The solution is derived from LR coefficients of $\theta=0\degree$ and $\theta=90\degree$.~\nodeLabel{d} We filter the insignificant LR coefficients based on p-values. Features with insignificant p-values are not visualized.~\nodeLabel{e} The largest coefficient of each feature is visualized in the Feature Clock.}
\label{fig:teaser}

%% file: 1-Introduction.tex
Dimensionality reduction methods transform high-dimensional data into lower-dimensional space. 
These methods aim to preserve various properties of the original data in lower dimensions (\textit{e.g.}, variance, pair-wise distances or similarities, or grouping structure). 
Use cases include numerous applications: feature selection~\cite{Zhangyu23QY, Hongfang19ZY}, visualization~\cite{Virgilio22, Yates22VB, Prummer23}, compression~\cite{PCACompression, jaradat2021tutorial}, and approximate techniques to avoid the curse of dimensionality~\cite{indyk1998approximate, berisha2021digital}. 
There are two main types of dimensionality reduction: linear and nonlinear.

Linear dimensionality reduction (LDR) techniques linearly project higher dimensional data into a lower dimensional space.
All LDR methods can be seen as a single matrix multiplication, according to $\mat{Y} = \mat{X} \mat{W}$ where $\mat{X}$ is the original data with samples as rows and features as columns, $\mat{Y}$ is the low-dimensional representation, and $\mat{W}$ is the transformation matrix. 
One of the most common visualization techniques to show the effect of high-dimensional features in LDR space is a biplot~\cite{biplot}, which depicts the rows of $\mat{W}$.


Nonlinear dimensionality reduction (NLDR), also called manifold learning, is a set of techniques that aim to project high-dimensional data onto lower-dimensional manifolds~\cite{tSNE, UMAP, PHATE}.
NLDR algorithms are usually better at preserving data grouping structure in two dimensions than LDR; this is a desirable behavior found in the outputs of many NLDR methods~\cite{tSNE, wattenberg2016how, UMAP, SainburgML2020, PHATE}.
A classical example of NLDR is t-distributed stochastic neighbor embedding (t-SNE)~\cite{tSNE, wattenberg2016how}, which constructs two probability distributions over high- and low-dimensional data and minimizes the Kullback–Leibler divergence (KLD)~\cite{KL} between the two distributions.
A disadvantage of NLDR is that the distances in the low-dimensional space may not reflect the distances in the original space~\cite{Linderman2017ClusteringWT}. 
Constructing biplots for NLDR is impossible because no linear $\mat{W}$ exists to explain the effect of features. 
Currently, one can visualize the effect of each feature of interest in a separate plot~\cite{Yates22VB, Virgilio22, ManduchiMarcinkevics2022}. 
These numerous plots are one of the best methods to understand the original feature's effects in low-dimensional spaces, but this solution is not scalable.

This paper introduces three types of static visualizations, highlighting the contributions of the high-dimensional features to linear directions of the two-dimensional (2D) spaces produced by NLDR.
The three techniques are a Global Feature Clock indicating the direction of features' contributions in low-dimensional space for the whole dataset, a Local Clock explaining features' impact within selected points, and an Inter-group Clock visualizing contributions between groups of points. 
The implementation is an open-source Python package.
Our technical contributions include: (1) Feature Clock, a novel technique for plotting feature contributions using linear regression in~\cref{Methods}, (2) a formal proof that ensures a correct behavior of Feature Clock in~\cref{Methods} and Suppl. Materials, and
(3) an experimental evaluation of the proposed visualization technique in several application cases in~\cref{results}. 



%% file: 2-background.tex
\section{Background and Related Work}
\label{sec:background}


This section summarizes common approaches to visualize the results of LDR and NLDR methods and highlights their limitations.

\begin{figure}[!t]
    \begin{tikzpicture}
        \node{
            \includegraphics[width=1\linewidth]{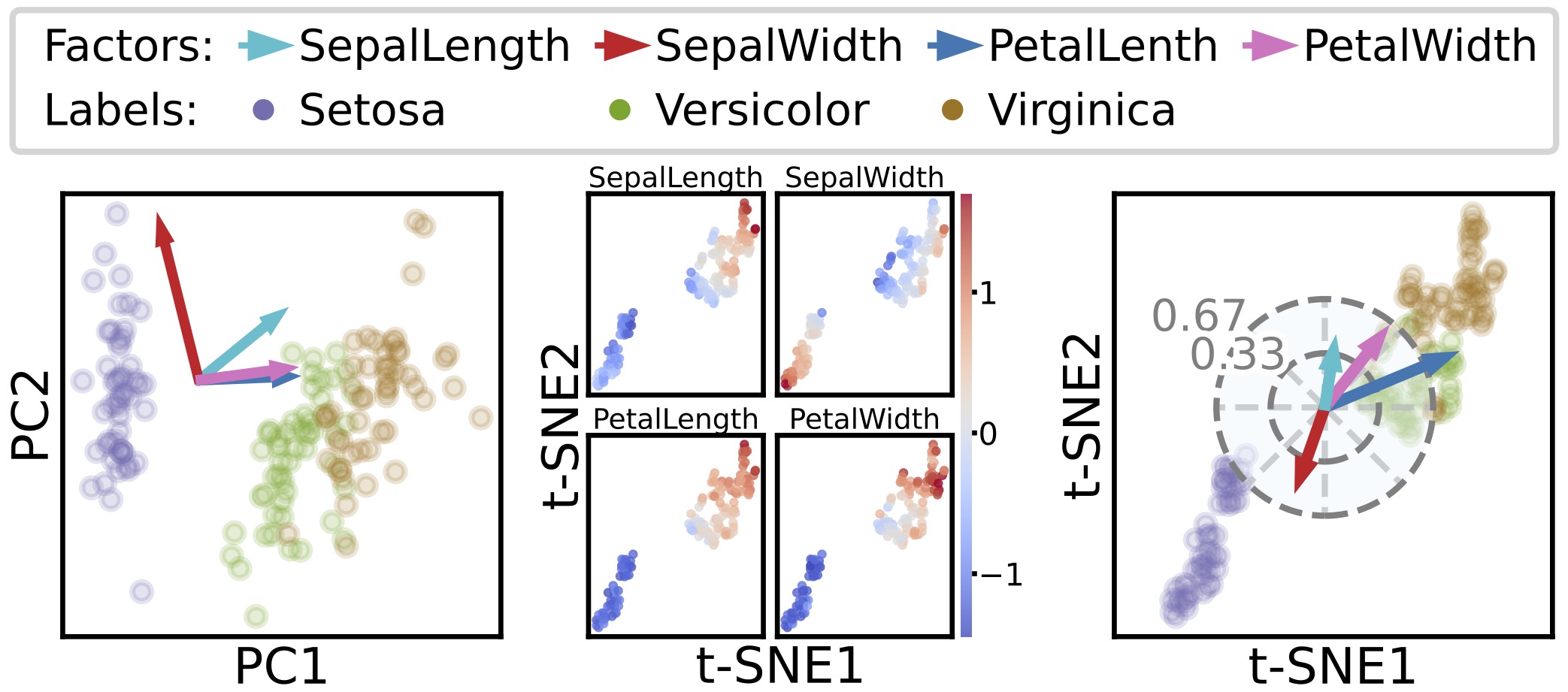}
        };
        \node at (-4.1cm,0.8cm){{a}};
        \node at (-1.23cm,0.8cm){{b}};
        \definecolor{colorsp}{HTML}{00c0ce}
        \definecolor{colorsw}{HTML}{d52728}
        \definecolor{colorpl}{HTML}{1f77b5}
        \definecolor{colorpw}{HTML}{e177c2}
        \node[fill=colorsp, draw=black, minimum height=4, minimum width=4, inner sep=0.0] at (-0.98cm,0.75cm){};
        \node[fill=colorsw, draw=black, minimum height=4, minimum width=4, inner sep=0.0] at (0.04cm,0.75cm){};
        \node[fill=colorpl, draw=black, minimum height=4, minimum width=4, inner sep=0.0] at (-0.98cm,-0.55cm){};
        \node[fill=colorpw, draw=black, minimum height=4, minimum width=4, inner sep=0.0] at (0.04cm,-0.55cm){};
        \node at ( 1.6cm,0.8cm){{c}};
    \end{tikzpicture}
\vspace{-0.7cm}
    \caption{(a) PCA biplot with feature loadings, (b) t-SNE scatter plots for each high-dimensional feature (perplexity=17). (c) A Feature Clock showing the feature contributions for the Iris dataset.}
    \label{fig:biplot}
    \vspace{-0.533cm}
\end{figure}

\textbf{Biplot for LDR:}
Biplot~\cite{biplot} is a visualization technique that can be applied to all LDR techniques. It involves creating a scatter plot that represents the low-dimensional data points, called a \emph{score plot}. 
Vectors are depicted to show the strength of each feature influence (rows of $\mat{W}$), which are referred to as a  \emph{loading plot}.
By interpreting a biplot, a user extracts information about the direction and strength of the association between original and low-dimensional space features.
The effect of $\mat{W}$ is uniform across the low-dimensional space, making biplot an effective tool.
\cref{fig:biplot}a shows a PCA biplot for the Iris flower dataset~\cite{IrisData} where the respective x- and y-axis are the two principal components. 
The points are 2D embeddings of the four-dimensional data points representing realizations of the iris plants. 
The arrows point toward change for a given feature. 
For instance, sepal width points at  {\raise.17ex\hbox{$\scriptstyle\mathtt{\sim}$}}$100\degree$, meaning that increasing sepal width translates to moving points up at the same angle.
A change in petal length and width affects embedded coordinates in the same direction. 
The magnitude of an arrow indicates how significant a change of the original feature affects shifts in coordinates. 



\textbf{Visualizations of feature effects for NLDR: }
While NLDR is popular for data visualization, interpreting the global and local structures of NLDR outputs can be challenging. 
A common method to assess the retained information in the low-dimensional representation involves assigning color codes to each projected data point based on a feature from the input data~\cite{Liu2017Visualizing}, which often results in visual clutter. Thus, field-based techniques can be used to reconstruct a 2D scalar function for a feature of interest used for spatial color coding~\cite{Sohns2022Attribute}. 
Other approaches illustrate regions of maximum attribute values. For example, DataContextMap~\cite{Cheng2016TheDC} enhances the projection by adding extra data points representing high feature values and displaying contours to depict feature distribution. DimReader~\cite{Faust2019DimReader} uses perturbations, measures their effect on the projection's outcome, and visualizes the resulting changes as grid lines. t-viSNE~\cite{tviSNE} supports exploring t-SNE projections, employing visualizations to analyze correlations between dimensions and visual patterns.

Many methods aim to identify clusters in the 2D space and explore their properties in the high-dimensional data. For instance, Joia et al.~\cite{Joia2015Uncovering} propose a visualization that utilizes matrix decomposition to define clusters in the visual space. 
Turkay et al.~\cite{Turkay2012Representative} associate representative factors with groups of dimensions for iterative analysis of the projected data, presenting techniques to compare and evaluate these factors.
Several approaches focus on comparing the properties of different groups found in the 2D space. For example, Probing Projections ~\cite{Stahnke2016Probing} enable to examine projections based on approximation errors and the influence of data dimensions on the projection space, and supports the comparison of different clusters in the 2D space. Eckelt et al.~\cite{Eckelt2023Visual} introduce new designs for visual summaries of common properties of such clusters and propose novel group comparison visualizations. Marcilio-Jr and Eler~\cite{MarcilioJr2021_ClusterShapley} employ cluster-oriented analysis to explain dimensionality reduction techniques with the so-called ClusterShapley method. 
Fujiwara et al.~\cite{Fujiwara2020Supporting} present a visual analytics method that explains features’ contributions to the projected data by calculating each feature’s relative contribution to the contrast between one cluster and other clusters.


Despite the work on how to visualize NLDR embeddings, there is no biplot alternative for NLDR.
A common solution is to show a plot per feature to analyze high-dimensional values in the low-dimensional space, see~\cref{fig:biplot}b.
Multiple scatter plots are a space-consuming and not scalable solution.
We suggest an alternative to the biplot: Feature Clock shown in~\cref{fig:biplot}c and described in \cref{sec:feature-clock}. 

%% file: 3-methods.tex
\section{Methodology}
\label{Methods}

This section outlines the internals of the Feature Clock, see~\cref{fig:teaser}.
The Feature Clock helps explore feature contributions of data points projected in two dimensions.
A user can be interested in how high-dimensional features contribute to the positions of low-dimensional data points globally or locally, or which features discriminate user-defined classes or clusters.
To address the former, we provide the Global and Local Clocks where the arrows correspond to the largest coefficients of linear regression, predicting coordinates of data points projected on a line passing through the center(s) of mass.
For the latter, we offer the Inter-group Clock, where arrows correspond to the feature contributions in the classification task (logistic regression).

\textbf{Feature Clock Overview:}\label{sec:feature-clock}
The method addresses a linear regression problem with the high-dimensional data as the predictor variable, $\textbf{X} \in \mathbb{R}^{n \times d}$, and the low-dimensional data, $\textbf{Y} \in \mathbb{R}^{n \times 2}$, projected on a line at angle $\theta$ as the target variable, $\mat{y}_\theta \in \mathbb{R}^{n}$ (\cref{fig:semicircleProjection}). 
We solve the optimization problem $\theta^j=\argmax_\theta{\lvert\beta^j\rvert}$ for each feature $j$, by finding the angle $\theta^j$, at which the absolute value of linear regression coefficient $\beta_{\theta}^j$ is maximized.
For each feature $j$, a closed-form solution allows to find the largest coefficient using coefficients $\beta_{0\degree}^j$, $\beta_{90\degree}^j$, and Pythagoras theorem (\cref{fig:teaser}c).
First, the high- and low-dimensional data are optionally normalized
(default: standardizing high-dimensional input, centering low-dimensional coordinates).
Second, the angle and magnitude of the strongest contribution are derived from the linear regression coefficients $\beta_{0\degree}$ and $\beta_{90\degree}$
(a formal proof in Sec.~A.1 of Suppl. Materials).
We fit two multivariate linear regression models to find the contribution of high-dimensional features $\mat{X}$ to the low-dimensional projections on the $x$ ($y_{0\degree}$) and $y$ ($y_{90\degree}$) axes in the 2D space.
To compute $\beta_{0\degree}$ and $\beta_{90\degree}$, one model is fitted between $X$ and $y_{0\degree}$, another between $X$ and $y_{90\degree}$.
We advise the user to standardize $\textbf{X}$ to make coefficients $\beta^j$ comparable across features; alternatively, we offer an option to standardize $\beta^j$, as \textit{post-processing}.
For each feature $j$, we define the biggest contribution as 
$(\beta_{\theta^j}^j)^2=(\beta_{0\degree}^j)^2 + (\beta_{90\degree}^j)^2$ 
at angle $\theta^j=\arctan(\beta_{90\degree}^j / \beta_{0\degree}^j)$.
Finally, we use t-test p-values to filter out statistically non-significant coefficients and visualize only significant contributions.
The p-value of a linear regression model checks if there is a significant linear relationship between each feature of $X$ and $y_{\theta}$. If the p-value is low, the relationship is significant (default: p-value $< 0.05$).


\input{figures/semicircleTikz}

\begin{figure*}[!t]
    \vspace{-0.3cm}
    \centering\begin{tikzpicture}
        \node{
            \includegraphics[width=\linewidth]{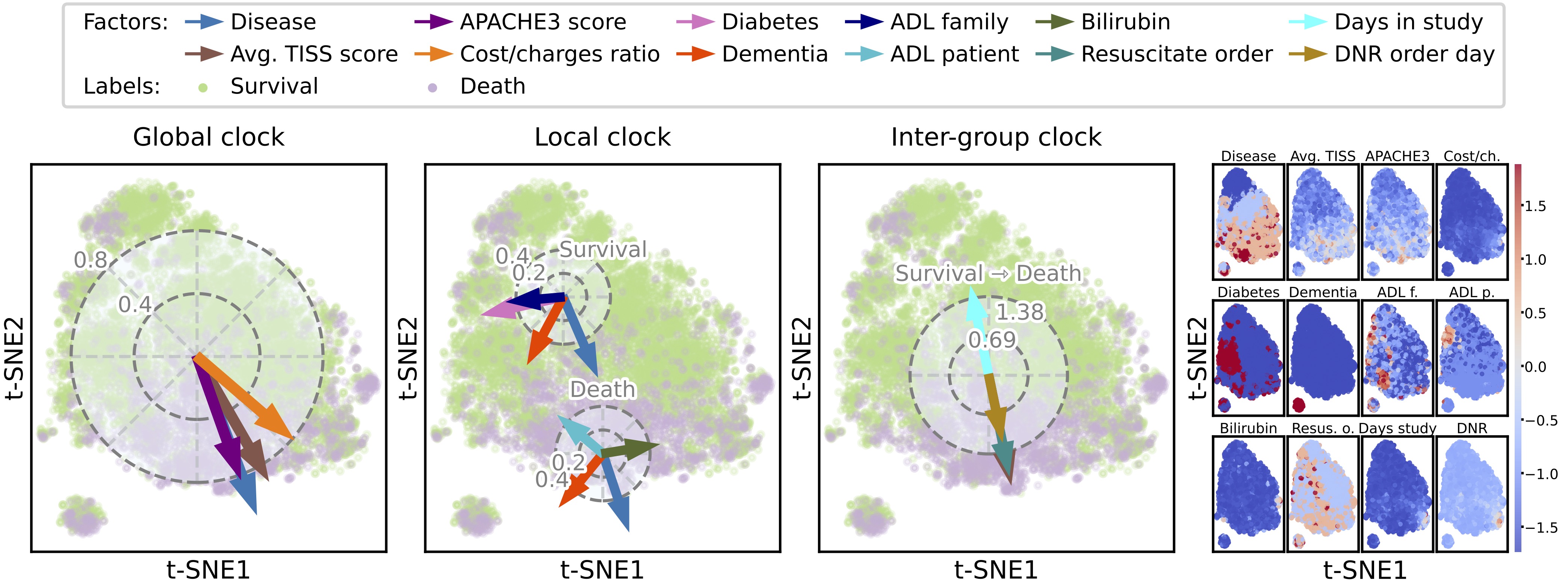}
        };
        \node at (-8.72cm,1.3cm){{a}};
        \node at (-4.23cm,1.3cm){{b}};
        \node at (0.26cm,1.3cm){{c}};
        \node at (4.73cm,1.3cm){{d}};

        \definecolor{disease}{HTML}{1f77b5}
        \definecolor{tiss}{HTML}{8c5749}
        \definecolor{apache}{HTML}{800180}
        \definecolor{cost}{HTML}{ff7f0b}

        \definecolor{diabetes}{HTML}{e177c2}
        \definecolor{dementia}{HTML}{fe4600}
        \definecolor{adlf}{HTML}{000081}
        \definecolor{adlp}{HTML}{27bdce}
        
        \definecolor{bili}{HTML}{556a2d}
        \definecolor{reso}{HTML}{018b8b}
        \definecolor{dstudy}{HTML}{12ffff}
        \definecolor{dnr}{HTML}{b68613}
        
        \node[fill=disease, draw=black, minimum height=4, minimum width=4, inner sep=0.0] at (5.59cm,1.4cm){};
        \node[fill=tiss, draw=black, minimum height=4, minimum width=4, inner sep=0.0] at (6.44cm,1.4cm){};
        \node[fill=apache, draw=black, minimum height=4, minimum width=4, inner sep=0.0] at (7.29cm,1.4cm){};
        \node[fill=cost, draw=black, minimum height=4, minimum width=4, inner sep=0.0] at (8.13cm,1.4cm){};

        \node[fill=diabetes, draw=black, minimum height=4, minimum width=4, inner sep=0.0] at (5.59cm,-0.15cm){};
        \node[fill=dementia, draw=black, minimum height=4, minimum width=4, inner sep=0.0] at (6.44cm,-0.15cm){};
        \node[fill=adlf, draw=black, minimum height=4, minimum width=4, inner sep=0.0] at (7.29cm,-0.15cm){};
        \node[fill=adlp, draw=black, minimum height=4, minimum width=4, inner sep=0.0] at (8.13cm,-0.15cm){};

        \node[fill=bili, draw=black, minimum height=4, minimum width=4, inner sep=0.0] at (5.59cm,-1.7cm){};
        \node[fill=reso, draw=black, minimum height=4, minimum width=4, inner sep=0.0] at (6.44cm,-1.7cm){};
        \node[fill=dstudy, draw=black, minimum height=4, minimum width=4, inner sep=0.0] at (7.29cm,-1.7cm){};
        \node[fill=dnr, draw=black, minimum height=4, minimum width=4, inner sep=0.0] at (8.13cm,-1.7cm){};
    \end{tikzpicture}
    \vspace{-0.7cm}
    \caption{(a) Global Feature Clock shows the general trend for top-4 significant features for hospital survival. (b) Local Clocks help to inspect the data within original labels (patient survived/died in hospital) while (c) Inter-group Clock shows how features change between two classes. (d) t-SNE (perplexity=50) scatter plots support three clocks and show the value range of high-dimensional features for survival in hospital.}
    \label{fig:supportClocks}
    \vspace{-0.5cm}
\end{figure*}

An alternative to visualizing one coefficient per variable is to project low-dimensional data on the radius of the semicircle at different angles, see~\cref{fig:semicircleProjection}, and visualize all coefficients for all features per projection, see Sec.~A.2 of Suppl. Materials. 
This results in perfect circles that appear because of projection and rotation.
Projection angles are defined by increments from $0\degree$ to $180\degree$ by a user-defined $\theta=\frac{180\degree}{m}$ with $m$ projection lines, where $\theta=5\degree$ is the default value. 
We default to plot the largest coefficient for each high-dimensional feature since it is a cleaner visualization that maintains maximum information, but the user can opt for the ``circles'' visualization.
If a dataset has many features, a user can visualize only top-$k$ features sorted by impact.

Clock annotations show the coefficient $\beta_{\theta}$ size and are intended for an easier comparison within and between clocks.
Comparison is possible because all coefficients are normalized.
We place the clock in the center of the low-dimensional points, but the user can change the location.
The clock location does not influence the computation of the contributions.
The size of the clock is defined by the distance between min and max points, but the user can scale the clock.

\textbf{Global Feature Clock:} 
We construct a clock using all data points to visualize the impact of the high-dimensional features in the two-dimensional space.
The disadvantage of the Global Clock is that it does not capture finer changes of gradients across the whole data, but rather a trend. 
Since we approximate the path on a manifold with a line, this clock can miss some information.

\textbf{Local Feature Clock:}
We use a Local Clock technique to explore data at a finer granularity within a group. 
The Local Clocks enable an easier analysis of a selection of neighboring points. 
Class labels or unsupervised clustering of high-dimensional data determine the points used for a single clock.
Analyzing original labels gives a perspective on what drives low-dimensional data point coordinates within a particular class or cluster. 
We default clustering to HDBSCAN with user-chosen parameters that also identifies outliers not assigned to any cluster.
Users are responsible for the meaningful clusters and can define their own clusters.
For each cluster or class, we apply the same method as in~\cref{fig:teaser} and create a single clock signifying changes within selected points.

\textbf{Inter-group Feature Clock:}
The third visualization option helps inspect how variables change between groups, either clusters or user-defined classes.
We fit a binary logistic regression with high-dimensional observations as predictor variables and visualize statistically significant coefficients on a single line that connects the group centers.
The arrow lengths correspond to the absolute values of the model coefficients. 
The clock is placed in the center of the line that connects the groups' centers.
In a multi-group setting, there is a space limitation for plotting all pairs of groups and their clocks.
Therefore, we build a minimum spanning tree (MST) between group centers in low-dimensional space and plot the inter-group clocks only for a trajectory on the MST.

\textbf{Limitations:} 
We explain high-dimensional contributions with linear directions.
If data is dense, we can approximate the movement of the gradients on the manifold in the original high-dimensional space with the shortest distance on average in low-dimensional space~\cite{Cheng2013HW}.
We assume that the shortest path is a line. The NLDR maps tend to plot points sparsely located away from cluster centers relatively close to those centers due to the sigmoid shape of the joint probability functions and the form of the objective function (KLD or cross-entropy)~\cite{tSNE, UMAP}. As a result, these points have a minimal impact on the values of the regression coefficients $\beta^j$.
The Feature Clock technique, especially the Global Clock, can miss information and 
not always follow the actual manifold and a nonlinear path on it. 

\textbf{Example:}~\textit{\cref{fig:biplot} shows a PCA biplot with feature loadings, t-SNE scatter plots for each variable, and Feature Clock for the Iris dataset~\cite{IrisData}.
A biplot (\cref{fig:biplot}a) shows how features influence low-dimensional representation.
We can see that petal width and length are the driving factors for the versicolor and verginica classes.
The scatter plots, in~\cref{fig:biplot}b, display how high-dimensional values of each feature are distributed in the 2D t-SNE space.
All features except sepal width increase the most with increasing x and y values.
From a Feature Clock in~\cref{fig:biplot}c, we observe how each variable impacts t-SNE embedding.
In the Feature Clock, the sepal length, petal width and length are driving factors for versicolor and virginica classes.
Sepal width increases in the direction of the setosa class, similar to the PCA.
For PCA, the Feature Clock produces the same arrows as the biplot (figures not shown). 
}

%% file: figures/semicircleTikz.tex
\setlength{\columnsep}{0.4cm}%
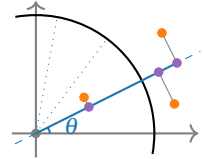
\begin{wrapfigure}{l}{3.3cm}
    \centering
    \vspace{-0.3cm}
    \begin{tikzpicture}[scale=0.8]
        \draw[gray, thick, ->] (-0.4,0) -- (2.7,0);
        \draw[gray, thick, ->] (0,-0.4) -- (0,2.2);

        \draw[thick] (1.967739,0) arc(0:100:1.967739);
        \draw[thick] (1.967739,0) arc(0:-10:1.967739);

        \filldraw[gray] (0,0) circle (2pt);

        \draw[mplBlue, thick, -](0,0) -- (3/1.25,1.5/1.25);
        \draw[mplBlue, dashed](-0.697305/2,-0.348652656/2) -- (3.3610208/1.23,1.68051/1.23);
        \draw[mplBlue, dotted, -](0,0) -- (1.180438921199,1.57419185);
        \draw[mplBlue, dotted, -](0,0) -- (0.352,1.936);
        
        \draw[mplBlue, thick, ]
            (0.2296,0) .. controls (0.2277,0.0519)..
            (0.1890,0.0945);
        \node[anchor=west, mplBlue] at(0.3277,0.119){$\theta$};

        \draw[gray] (0.8,0.6)            -- (0.88,0.44);
        \draw[gray] (2.301929,0.4901)    -- (2.040349,1.0201746);
        \draw[gray] (2.0942574,1.670811) -- (2.34373,1.171865);
        
        \filldraw[mplOrange] (0.8,0.6) circle (2pt);
        \filldraw[mplOrange] (2.301929,0.4901) circle (2pt);
        \filldraw[mplOrange] (2.0942574,1.670811) circle (2pt);

        \filldraw[mplPurple] (0.88,0.44) circle (2pt);
        \filldraw[mplPurple] (2.040349,1.0201746) circle (2pt);
        \filldraw[mplPurple] (2.34373,1.171865) circle (2pt);

    \end{tikzpicture}

    \vspace{-0.2cm}
    \caption{{\color{mplPurple}Projection} \color{black} of {\color{mplOrange}2D points} on a line at angle $\theta$.}
    \vspace{-0.3cm}
    \label{fig:semicircleProjection}
\end{wrapfigure}

%% file: 4-evaluation.tex
\section{Use Cases}
\label{results}

In this section, we describe several applications of Feature Clocks and how they improve explainability for each task.


\subsection{Analysis of Low-dimensional Data}

Support2~\cite{support2} is a dataset that comprises information on critically ill patients across US medical centers.
Each patient is assigned one of nine diseases: acute respiratory failure, liver disease, colon/lung cancer, etc. 
The dataset is often used to train supervised machine learning models to determine whether patients would die in the hospital based on physiological, demographic, and disease information. 

There are 46 features describing each patient. 
It is infeasible to visualize the effects of all features by plotting 46 scatter plots, one for each variable.
Therefore, the high-dimensional feature's impact can be visualized by Feature Clocks with the top-$k$ significant features.
\cref{fig:supportClocks} shows three clocks and scatter plots supporting clocks for the Support2 data. For all three clocks, we have set $k=4$ (top-$4$ features). 
We use the hospital death status of the patient as a label for the Local and Inter-group Clocks. 
We observe that t-SNE captures the difference between patients who die or survive in the hospital.

The Global Clock (\cref{fig:supportClocks}a) shows that disease class, cost/charges ratio, average Therapeutic Intervention Scoring System (TISS), and APACHE3 scores differentiate the outcome of critically ill patients.
To support the Global Clock, we show scatter plots in~\cref{fig:supportClocks}d.
We see that patients with a less severe disease class survive more often, and medical centers charge these patients less.
The average TISS score allows a quantitative comparison of patient care, research experiences of different intensive care units (ICU), and an estimate of the severity of the disease.
Most of the deceased patients had a high TISS score. 
APACHE3 scoring serves as an early warning indication of death and prompts clinicians to upgrade the treatment protocol, making it a useful tool for the clinical prediction of ICU mortality.
The people with a negative outcome have a higher APACHE3 score.

The Local Clocks in~\cref{fig:supportClocks}b represent points that belong to either the survival or death class.
We see two clocks; both are driven by the disease class and the presence of dementia.
The 2D space of living patients is influenced by variables such as diabetes and the Activities of Daily Living (ADL) score estimated by the family of a sick person. 
The ADL score evaluates a patient's ability to accomplish their daily activities. 
The patient's ADL and bilirubin levels are significant contributors to the 2D representation of the patients who die.

In~\cref{fig:supportClocks}c, the Inter-group Clock shows features that differ between deceased and living patients.
As shown by the Inter-group Clock, mortality increases with higher values of the average TISS, Do Not Resuscitate (DNR) order day, and if resuscitation was given. 
Survival decreases with days spent in the study.

\begin{figure}[!t]
  \centering
  \begin{tikzpicture}
        \node{
            \includegraphics[width=1\linewidth]{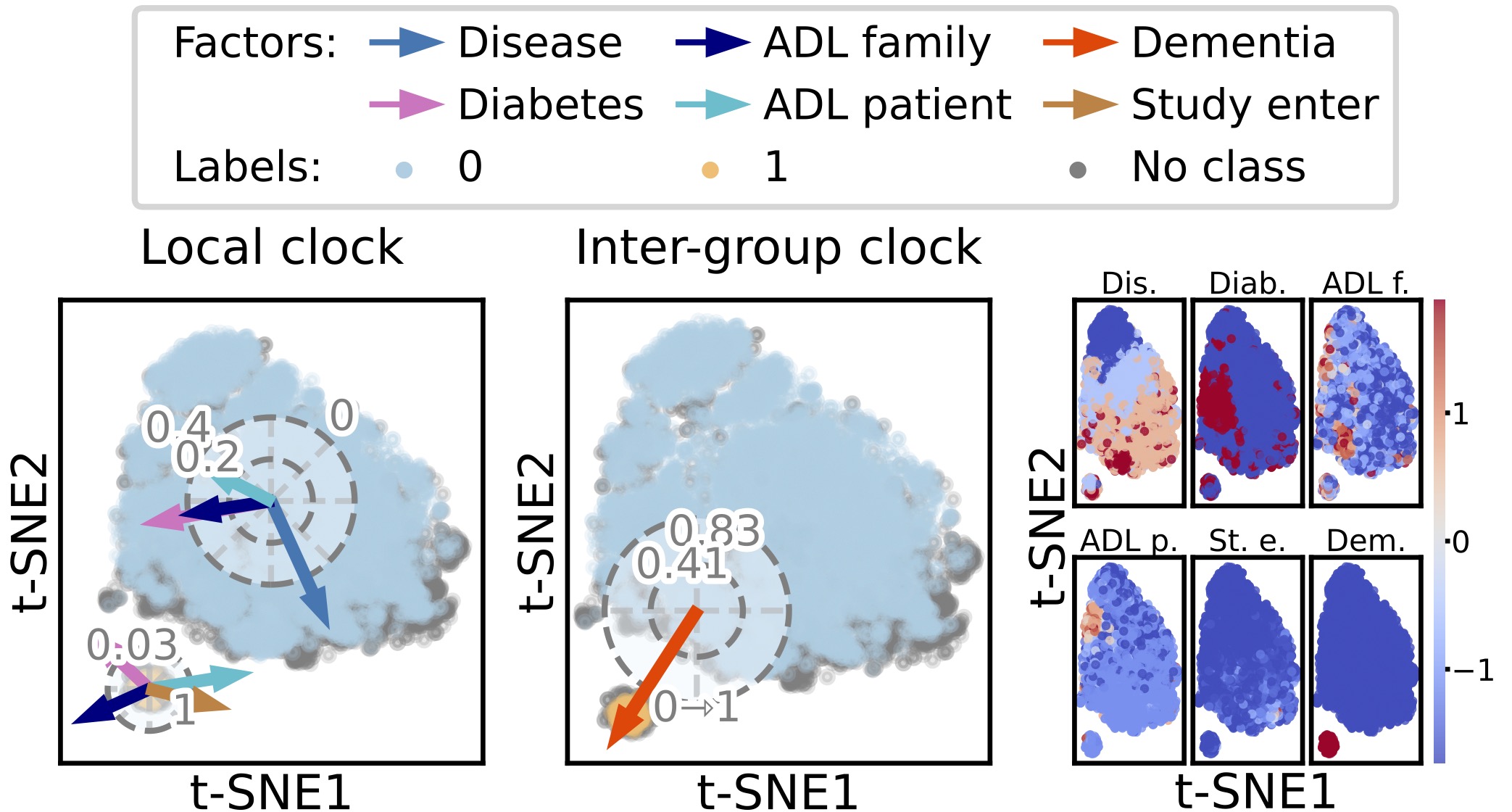}
        };
        \node at (-4.1cm,0.5cm){{a}};
        \node at (-1.23cm,0.5cm){{b}};
        \node at ( 1.6cm,0.5cm){{c}};

        \definecolor{disease}{HTML}{1f77b5}
        \definecolor{tiss}{HTML}{8c5749}
        \definecolor{apache}{HTML}{800180}
        \definecolor{cost}{HTML}{ff7f0b}

        \definecolor{diabetes}{HTML}{e177c2}
        \definecolor{dementia}{HTML}{fe4600}
        \definecolor{adlf}{HTML}{000081}
        \definecolor{adlp}{HTML}{27bdce}
        
        \definecolor{bili}{HTML}{556a2d}
        \definecolor{reso}{HTML}{018b8b}
        \definecolor{dstudy}{HTML}{12ffff}
        \definecolor{dnr}{HTML}{b68613}
        \definecolor{ste}{HTML}{cd853f}
        
        \node[fill=disease, draw=black, minimum height=4, minimum width=4, inner sep=0.0] at (2.32cm,0.52cm){};
        \node[fill=diabetes, draw=black, minimum height=4, minimum width=4, inner sep=0.0] at (2.98cm,0.52cm){};
        \node[fill=adlf, draw=black, minimum height=4, minimum width=4, inner sep=0.0] at (3.65cm,0.52cm){};

        \node[fill=adlp, draw=black, minimum height=4, minimum width=4, inner sep=0.0] at (2.32cm,-0.92cm){};
        \node[fill=ste, draw=black, minimum height=4, minimum width=4, inner sep=0.0] at (2.98cm,-0.92cm){};
        \node[fill=dementia, draw=black, minimum height=4, minimum width=4, inner sep=0.0] at (3.65cm,-0.92cm){};
    \end{tikzpicture}
  
  \vspace{-0.3cm}
  \caption{(a) Local Clocks explore the NLDR space by inspecting HDBSCAN clusters. (b) Inter-group Clock shows how features change between clusters. (c) For the verification, we show a t-SNE scatter plot for each feature in the clocks (perplexity=50).}
  \label{fig:support_hdbsacn}
  \vspace{-0.6cm}
\end{figure}

Instead of visualizing clocks for the original target, the user can explore low-dimensional space with clustering.
In~\cref{fig:support_hdbsacn}, the Local and Inter-group Clocks (top-$4$ features) use HDBSCAN clusters of high-dimensional data; the resulting two clusters are clearly seen in the t-SNE embedding (blue and orange). Some points are not assigned to any cluster by HDBSCAN (gray).
The Local Clock in~\cref{fig:support_hdbsacn}a shows that coordinates of data points in both clusters 0 and 1 are driven by the presence of diabetes and ADL scores.
Interestingly, the ADL score estimated by the patient's family is high on the left in the two clusters, and contributions point in the respective direction.
The same applies to the diabetes status.
A bigger cluster 0 is also characterized by disease class while cluster 1 differs by the day of the study entrance.

The Inter-group Clock in~\cref{fig:support_hdbsacn}b shows the feature that drives the difference between the two clusters. In the t-SNE space, people with dementia are clearly separated from patients without the disease, and that is the only significant feature for the trained classifier. 

Further use cases are included in the Supplemental Material.

\subsection{Neural Network Interpretability}
\begin{figure}[!b]
  \centering
  \vspace{-0.6cm}
  \begin{tikzpicture}
        \node{
            \includegraphics[width=1\linewidth]{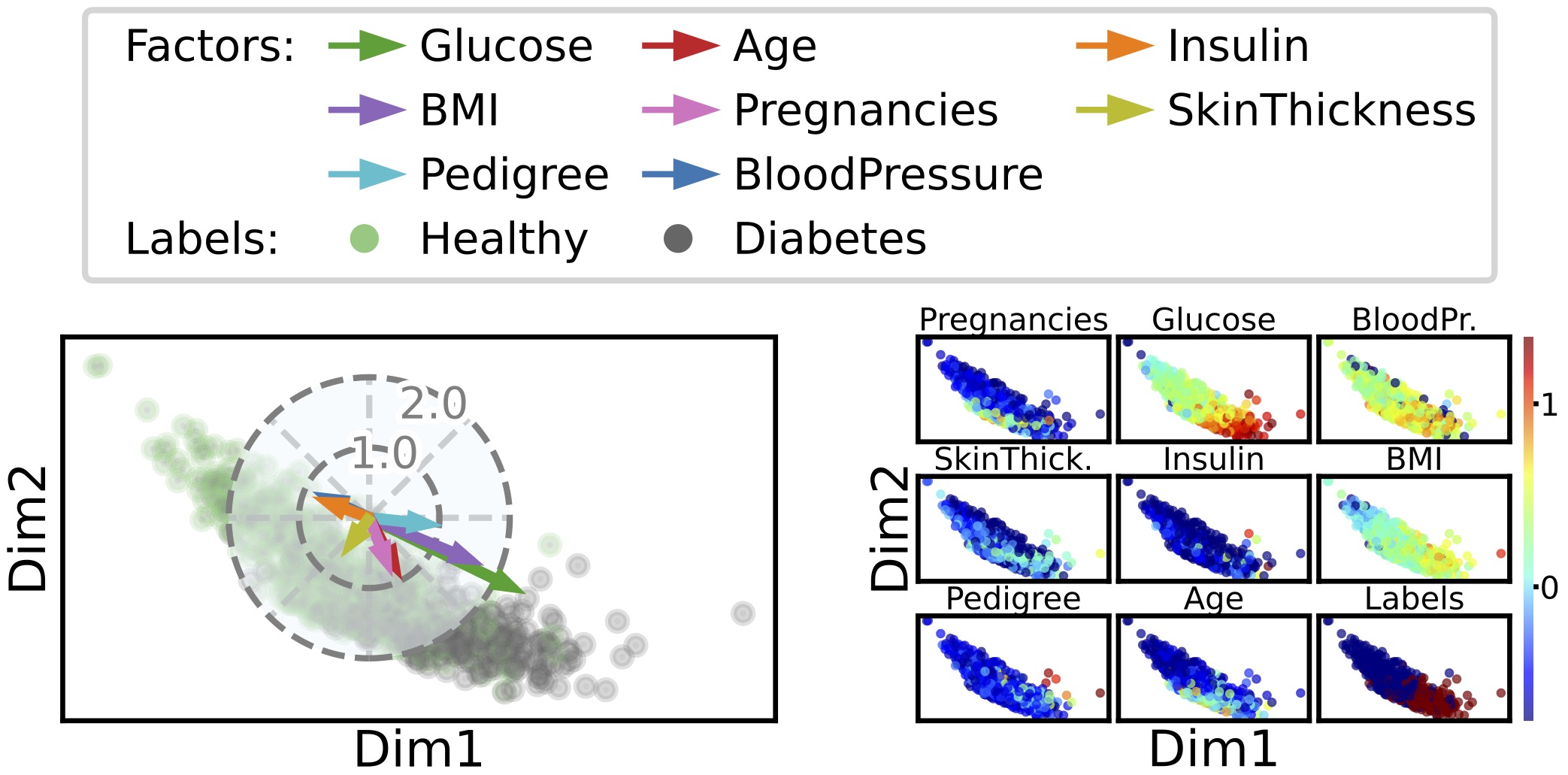}
        };
        \node at (-4.1cm,0.15cm){{a}};
        \node at (0.53cm,0.15cm){{b}};

        \definecolor{pr}{HTML}{e37ac3}
        \definecolor{glucose}{HTML}{2ca02c}
        \definecolor{blood}{HTML}{287cb7}
        \definecolor{skin}{HTML}{babe23}
        \definecolor{insulin}{HTML}{fe800b}
        \definecolor{bmi}{HTML}{9368b9}
        \definecolor{pedigree}{HTML}{1fbece}
        \definecolor{age}{HTML}{d32826}
        
        \node[fill=pr, draw=black, minimum height=4, minimum width=4, inner sep=0.0] at (1.68cm,0.2cm){};
        \node[fill=glucose, draw=black, minimum height=4, minimum width=4, inner sep=0.0] at (2.76cm,0.2cm){};
        \node[fill=blood, draw=black, minimum height=4, minimum width=4, inner sep=0.0] at (3.84cm,0.2cm){};

        \node[fill=skin, draw=black, minimum height=4, minimum width=4, inner sep=0.0] at (1.68cm,-0.55cm){};
        \node[fill=insulin, draw=black, minimum height=4, minimum width=4, inner sep=0.0] at (2.76cm,-0.55cm){};
        \node[fill=bmi, draw=black, minimum height=4, minimum width=4, inner sep=0.0] at (3.84cm,-0.55cm){};

        \node[fill=pedigree, draw=black, minimum height=4, minimum width=4, inner sep=0.0] at (1.68cm,-1.306cm){};
        \node[fill=age, draw=black, minimum height=4, minimum width=4, inner sep=0.0] at (2.76cm,-1.306cm){};
    \end{tikzpicture}
  
  \vspace{-0.3cm}
  \caption{(a) We use Global Clock to describe 2D hidden state of the NN and justify the clock with (b) a scatter plot for each feature.}
  \label{fig:pima_nn}
  \vspace{-0.3cm}
\end{figure}

The Pima Indians Diabetes Database~\cite{pimaData} provides data on adult females of Pima Indian heritage. 
It is used to train models predicting if a patient has diabetes based on clinical and demographic data.

The clock technique can be used to explain the hidden states of a neural network (NN). 
Users can plot a Feature Clock using either a 2D hidden state or apply NLDR to any high-dimensional hidden layer to get a 2D representation. This visualization has the potential to help the user better understand the factors that drive the information extracted at each hidden layer.

In this use case, we analyze the last hidden layer (two neurons) of a NN with two layers. 
\cref{fig:pima_nn} shows a clock and scatter plots for the last hidden layer, using true labels in the visualization.
The representations of patients with and without diabetes are almost linearly separable.
The clock on~\cref{fig:pima_nn} identifies increasing glucose and body mass index (BMI) as indicating factors for diabetes.
This conclusion can be verified by scatter plots in~\cref{fig:pima_nn}, where we show a subplot for each feature, including the target \textit{Labels}.
Age and number of pregnancies also substantially contribute to the activation values of this hidden layer.
Low insulin and blood pressure lead to no diabetes, which aligns with low values in the scatter plots.

It is worth noting that all Feature Clock versions are applicable to any NN architecture, including autoencoders. We used the feed-forward NN with two neurons as a simple example.

\subsection{Preliminary User Feedback}
We gathered feedback regarding the readability of the Feature Clock visualization in an informal interview with two post-doctoral researchers from the visual analytics research field. The researchers agreed that the visualization is simple yet effective, easy graspable, and intuitive. They also 
mentioned potential extensions in highlighting the uncertainty in the feature contribution through the visual blurring of the respective arrows as well as extending the visual representation to those used to represent velocity~\cite{Gorin2022RNA}.

%% file: 5-discussion.tex
\section{Reflections, Issues and Future Work}
To summarize, we have shown a group of novel plotting techniques called Feature Clocks.
We illustrate how Feature Clocks can be used to interpret lower-dimensional space resulting from NLDR methods and the latent space of deep neural networks.

The implication of the suggested technique is an intuitive, compact, and informative visualization that enhances explainability.
It eliminates the need for multiple scatter plots to explain the impact of original features after a non-linear dimensionality reduction.
For a large number of variables, making scatter plots for each feature becomes infeasible and challenging to analyze visually.
The main limitation is the linear approximation of the nonlinear path on the manifold. This path might not exist if manifolds are disconnected.
Feature Clock advises a user about the feature changes, but some information might be missed, especially in the Global Clock, since we might not follow the actual manifold.

Future work may include refinement of the model for the visualization.
One can use non-linear regression to visualize complex gradient changes in the feature space. The challenge lies in developing an algorithm that is computationally efficient enough to solve the optimization problem. Another potential direction of future research is replacing $\mathbf{y_\theta}$ with a trajectory, non-linear in 2D, better reflecting the shortest-distance traverse of the low-dimensional manifold.
Additionally, to allow interactive selection of the points to analyze, we could implement a lasso tool using GPU acceleration.


%% file: 6-conclusions.tex
\section{Conclusions}

We introduced Feature Clock, a group of novel plotting techniques for global, local, and inter-group analysis of non-linear two-dimensional spaces.
First, the Feature Clock enables compact representations of the two-dimensional space, highlighting contributions of high-dimensional features.
Second, we found an efficient way to identify the largest feature contribution: We mathematically proved the validity of calculating the biggest contribution from two projections and showed that expected behavior matches the empirical observations in the plots.
Third, we have illustrated that the Feature Clock is applicable to the analysis of the classical non-linear dimensional reduction and the neural networks' latent space.

%% file: 7-acknowledgements.tex
\acknowledgments{%
We are grateful to Sapar Charyyev for the comments on the mathematical formulation of the problem.
We thank Sebastian Baunsgaard and Florian Barkmann for their comments on the early drafts and for engaging in helpful discussions.
}

%% file: 8-appendix.tex
\appendix
\section{Supplementary Materials}
\label{appendix:all}

In this document, we present mathematical proof of the Feature Clock\footnote{\url{https://pypi.org/project/feature-clock}} and further applications of Global, Local and Inter-group Clocks.
Due to page constraints, we could not include them in the full size in the paper.

\subsection{Theoretical Foundation}\label{matproof}

\input{figures/circleTheoremTikz}

We want to formally prove that clock technique is valid, and resulting perfect circles are the expected behavior (see Figure~\cref{fig:circles}).
For simplicity, it is enough to show that transformation (projection and rotation) results in circles for one high-dimensional feature.

First, the low-dimensional data $X \in \mathbb{R}^{nx2}$ is projected on the radius of the semicircle at angles from $0\degree$ to $180\degree$ with $5\degree$ increments. 
For some projection angle $\theta$, $\cos(\theta) = \frac{d}{\left\| X \right\|} = \frac{X\cdot v}{\left\| X \right\| \left\| v \right\|}$ where $v=(\cos{\theta}, \sin{\theta})$ is a unit vector ($\left\| v \right\|=1$), and $d$ is the distance vector from the origin of the semicircle to the projected points.
It is obvious that $d=X\cdot v$, $d \in \mathbb{R}^{nx1}$, $X \in \mathbb{R}^{nx2}$, $v \in \mathbb{R}^{2x1}$.

Second, for each $\theta$, linear regression (LR) is used to predict projection $d$ from the high-dimensional data $S \in \mathbb{R}^{nxd}$, with a closed form solution $\beta_{\theta} \in \mathbb{R}^{dx1}$:
\begin{equation}
\beta_{\theta}=(S^TS)^{-1}S^Td=(S^TS)^{-1}S^TXv=
AX
\begin{bmatrix}
\cos{\theta}\\
\sin{\theta}
\end{bmatrix}
\label{eq:beta}
\end{equation}
where $A=(S^TS)^{-1}S^T$.
$\beta_{\theta}$ is a vector with each feature's LR coefficient for a projection at angle $\theta$.

It is known that three non-collinear points make a unique circle. 
It can be shown by constructing perpendicular bisectors through each of the line segments formed between three points that do not lie on the same line. 
All perpendicular bisectors intersect at one point and are all equidistant from that intersection point. Therefore, we get a circle.

We always project on $0\degree$ and $90\degree$ from origin.
Therefore, as shown in~\cref{fig:circleTheorem}, we choose three non-collinear points: Origin $O$, $B$ on the y-axis, and $D$ on the x-axis.
We need to prove that fourth point $C$ lies on the circumcircle of $O$, $B$, $D$ by showing that $\angle BCD=90\degree$. 
If~\cref{eq:angleequation} holds, then $\angle BCD=90\degree$, and $O$, $B$, $C$, $D$ form a circle.
\begin{equation}
{BD}^2 = {BC}^2 + {CD}^2 = {BO}^2 + {OD}^2    
\label{eq:angleequation}
\end{equation}

Let's choose a i-th row of $\beta_{\theta}$ and $A$, scalar $\beta_{\theta}^{i}$ is a coefficient of i-th feature for angle $\theta$ and $a_i=((S^TS)^{-1}S^T)_i$.
From~\cref{eq:beta},
$\beta_{0\degree}^i=a_iX\begin{bmatrix}
1\\
0
\end{bmatrix}
$
 and
$\beta_{90\degree}^i=a_iX\begin{bmatrix}
0\\
1
\end{bmatrix}
$ for $\theta=0\degree$ and $\theta=90\degree$ respectively.

We define four points 
$C=(\beta_{\theta}^i \cos{\theta}, \beta_{\theta}^i \sin{\theta})$,
$B=(0, \beta_{90\degree}^i)$,
$D=(\beta_{0\degree}^i , 0)$, and
$O=(0\degree, 0\degree)$, then ${OD}=\beta_{0\degree}^i$ and
${BO} = \beta_{90\degree}^i$.

\begin{equation}
  \begin{aligned}
    {OD}^2+{BO}^2 &= 
    a_iX\begin{bmatrix}
    1\\
    0
    \end{bmatrix}
    \begin{bmatrix}
    1&0\\
    \end{bmatrix}
    X^Ta_i^T
    + 
    a_iX\begin{bmatrix}
    0\\
    1
    \end{bmatrix}
    \begin{bmatrix}
    0&1\\
    \end{bmatrix}
    X^Ta_i^T \\ &=
    a_iX\begin{bmatrix}
    1&0\\
    0&0
    \end{bmatrix}
    X^Ta_i^T
    + 
    a_iX\begin{bmatrix}
    0&0\\
    0&1
    \end{bmatrix}
    X^Ta_i^T \\ &=
    a_iX\begin{bmatrix}
    1&0\\
    0&1
    \end{bmatrix}
    X^Ta_i^T
  \end{aligned}
  \label{eq:osquare}
\end{equation}

\begin{equation}
  \begin{aligned}
    {BC}^2 &= 
    (-\beta_{\theta}^i \cos{\theta})^2 + (\beta_{90\degree}^i-\beta_{\theta}^i \sin{\theta})^2 \\ &=
    \left( a_iX\begin{bmatrix}
    \cos{\theta}^2\\
    \sin{\theta}\cos{\theta}
    \end{bmatrix}\right)^2 + 
    \left(a_iX\begin{bmatrix}
    0\\
    1
    \end{bmatrix} - 
    a_iX\begin{bmatrix}
    \cos{\theta}\sin{\theta}\\
    \sin{\theta}^2
    \end{bmatrix}\right)^2 \\ &=
    \left( a_iX\begin{bmatrix}
    \cos{\theta}^2\\
    \sin{\theta}\cos{\theta}
    \end{bmatrix}\right)^2 + 
    \left(
    a_iX\begin{bmatrix}
    -\cos{\theta}\sin{\theta}\\
    \cos{\theta}^2
    \end{bmatrix}\right)^2 \\ &=
    \cos{\theta}^2
    \left( a_iX\begin{bmatrix}
    \cos{\theta}\\
    \sin{\theta}
    \end{bmatrix}\right)^2 + 
    \cos{\theta}^2
    \left(a_iX\begin{bmatrix}
    -\sin{\theta}\\
    \cos{\theta}
    \end{bmatrix}\right)^2 
  \end{aligned}
  \label{eq:bc}
\end{equation}

\begin{equation}
  \begin{aligned}
    {CD}^2 &= 
    (\beta_{0\degree}^i-\beta_{\theta}^i \cos{\theta})^2 + (-\beta_{\theta}^i \sin{\theta})^2\\ &=
    \left(a_iX\begin{bmatrix}
    1\\
    0
    \end{bmatrix} - 
    a_iX\begin{bmatrix}
    \cos{\theta}^2\\
    \sin{\theta}\cos{\theta}
    \end{bmatrix}\right)^2 +
    \left(a_iX\begin{bmatrix}
    \sin{\theta}\cos{\theta}\\
    \sin{\theta}^2
    \end{bmatrix}\right)^2 \\ &=
    \left( a_iX\begin{bmatrix}
    \sin{\theta}^2\\
    -\sin{\theta}\cos{\theta}
    \end{bmatrix}\right)^2 + 
    \left(
    a_iX\begin{bmatrix}
    \sin{\theta}\cos{\theta}\\
    \sin{\theta}^2
    \end{bmatrix}\right)^2 \\ &=
    \sin{\theta}^2
    \left( a_iX\begin{bmatrix}
    \sin{\theta}\\
    -\cos{\theta}
    \end{bmatrix}\right)^2 + 
    \sin{\theta}^2
    \left(a_iX\begin{bmatrix}
    \cos{\theta}\\
    \sin{\theta}
    \end{bmatrix}\right)^2
  \end{aligned}
  \label{eq:cd}
\end{equation}

We combine~\cref{eq:bc} and~\cref{eq:cd}, and $\sin{\theta}^2 + \cos{\theta}^2=1$:
\begin{equation}
  \begin{aligned}
    {BC}^2 + {CD}^2 &= 
    \left(a_iX\begin{bmatrix}
    \cos{\theta}\\
    \sin{\theta}
    \end{bmatrix}\right)^2 +
    \left( a_iX\begin{bmatrix}
    \sin{\theta}\\
    -\cos{\theta}
    \end{bmatrix}\right)^2 \\ &=
    a_iX\begin{bmatrix}
    \cos{\theta}\\
    \sin{\theta}
    \end{bmatrix}
    \begin{bmatrix}
    \cos{\theta}&\sin{\theta}\\
    \end{bmatrix}
    X^Ta_i^T
    \\ &+ 
    a_iX\begin{bmatrix}
    \sin{\theta}\\
    -\cos{\theta}
    \end{bmatrix}
    \begin{bmatrix}
    \sin{\theta}&-\cos{\theta}\\
    \end{bmatrix}
    X^Ta_i^T \\ &=
    a_iX\begin{bmatrix}
    \cos{\theta}^2&\sin{\theta}\cos{\theta}\\
    \sin{\theta}\cos{\theta}&\sin{\theta}^2
    \end{bmatrix}
    X^Ta_i^T
    \\ &+ 
    a_iX\begin{bmatrix}
    \sin{\theta}^2&-\sin{\theta}\cos{\theta}\\
    -\sin{\theta}\cos{\theta}&\cos{\theta}^2
    \end{bmatrix}
    X^Ta_i^T \\ &=
    a_iX\begin{bmatrix}
    1&0\\
    0&1
    \end{bmatrix}
    X^Ta_i^T 
  \end{aligned}
\label{eq:csquare}
\end{equation}

We can see that~\cref{eq:osquare} and~\cref{eq:csquare} are equal, $\angle BCD=90\degree$, point $C$ lies on the circumcircle, and, therefore, a circle is formed by projecting and rotating.

We know that $\angle BOD=90\degree$ and $BD$ is the diameter of a circle.
Since the biggest contribution $OK$ defines the circle, it is also the diameter of the circle.
The intersection of $OK$ and $BD$ is the center of a circle.
Therefore, we can find coordinates of $K$ after projecting and rotating on the x- and y-axis only, at angles of $0\degree$ and $90\degree$ respectively.
If $D=(x, 0)$ and $B=(0, y)$, then $K=(x, y)$.

\subsection{Visualization of all coefficients}

\begin{figure}[t]
    \centering
    \begin{subfigure}{0.49\linewidth}
    \includegraphics[width=\textwidth]{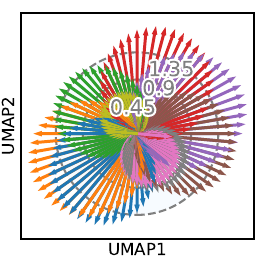}
        \caption{All}
        \label{fig:Circles_all}
    \end{subfigure}
    \hfill
    \begin{subfigure}{0.49\linewidth}
        \includegraphics[width=\textwidth]{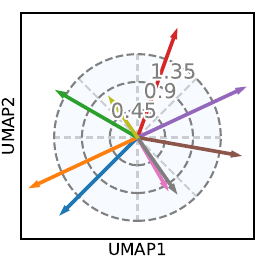}
        \caption{Largest}
        \label{fig:Circles_largest}
    \end{subfigure}

    \vspace{-0.3cm}
    \caption{All vs Largest Coefficients}
    \label{fig:circles}
\end{figure}

If we project on lines with $\theta$ iterates, we get perfect circles as shown in~\cref{fig:circles}.

\begin{figure*}[!t]
    \centering
    \begin{tikzpicture}
        \node{
            \includegraphics[width=0.9\linewidth]{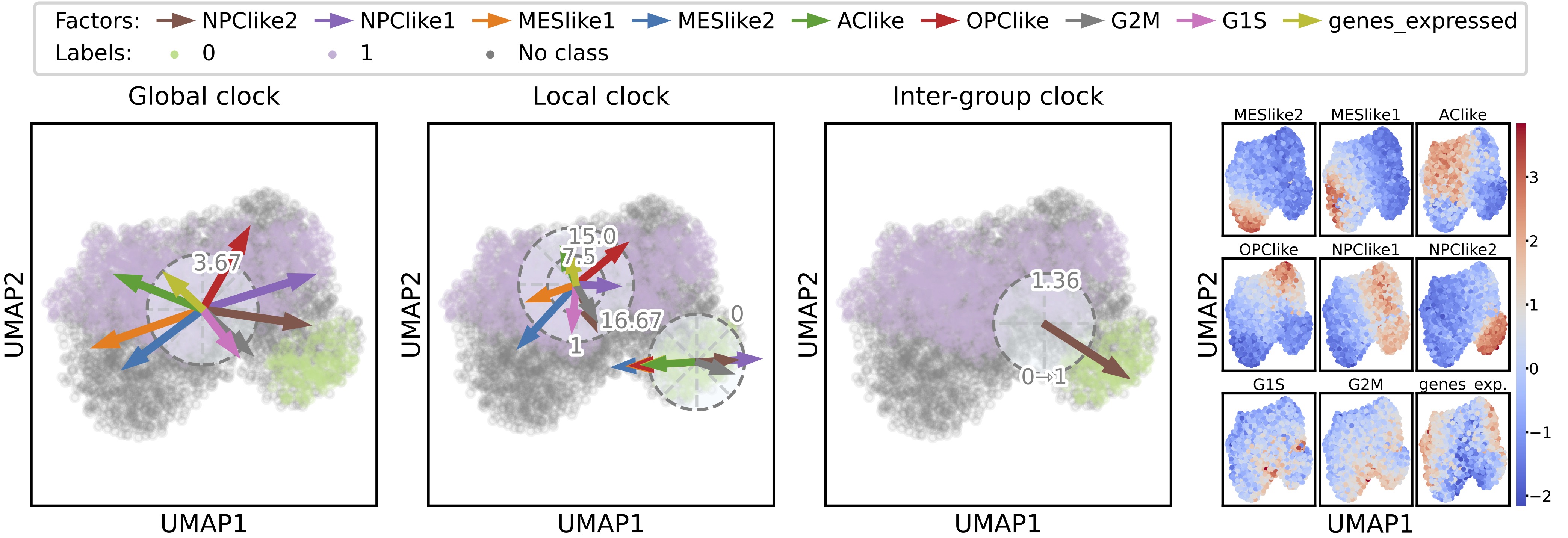}
        };
        \node at (-7.9cm,1.4cm){{a}};
        \node at (-3.83cm,1.4cm){{b}};
        \node at (0.25cm,1.4cm){{c}};
        \node at (4.3cm,1.4cm){{d}};
    \end{tikzpicture}
    \caption{(a)~The Global Feature Clock shows all significant glioblastoma signatures for all observations. (b)~Local Clocks analyze the data at the granularity of HDBSCAN clusters, and (c)~Inter-group Clock explains how features change between two clusters. (d)~Scatter plots help to validate Feature Clocks and visualize a plot per each high-dimensional feature. 
    \label{fig:neftelClocks}}
    \begin{tikzpicture}
        \node{
            \includegraphics[width=0.9\linewidth]{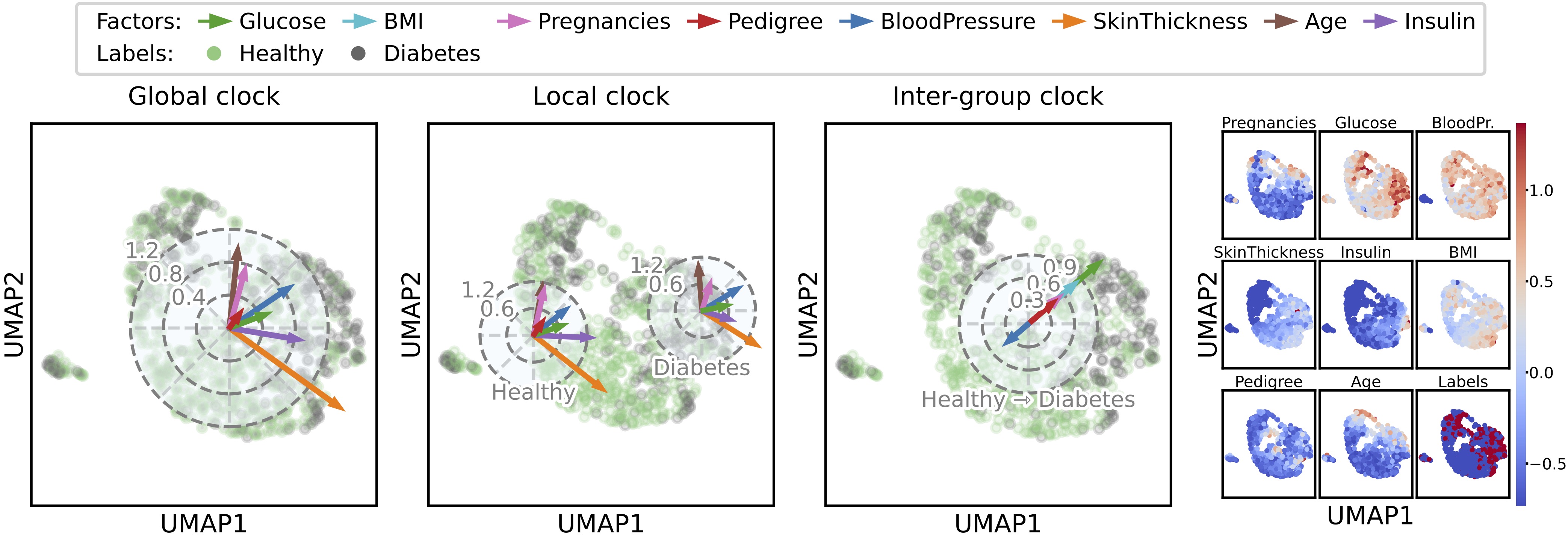}
        };
        \node at (-7.9cm,1.26cm){{a}};
        \node at (-3.83cm,1.26cm){{b}};
        \node at (0.25cm,1.26cm){{c}};
        \node at (4.3cm,1.26cm){{d}};a
    \end{tikzpicture}
    \caption{(a)~The Global Feature Clock highlights features that differentiate people with and without diabetes, while (b)~Local Clocks show how features range within original classes. (c)~Inter-group Clock describes how personal data differs between females with and without diabetes. (d)~Scatter plots are intended to support clocks and ease the users' perception of a new visualization.
    \label{fig:pimaClocks}}
    \begin{tikzpicture}
        \node{
            \includegraphics[width=0.9\linewidth]{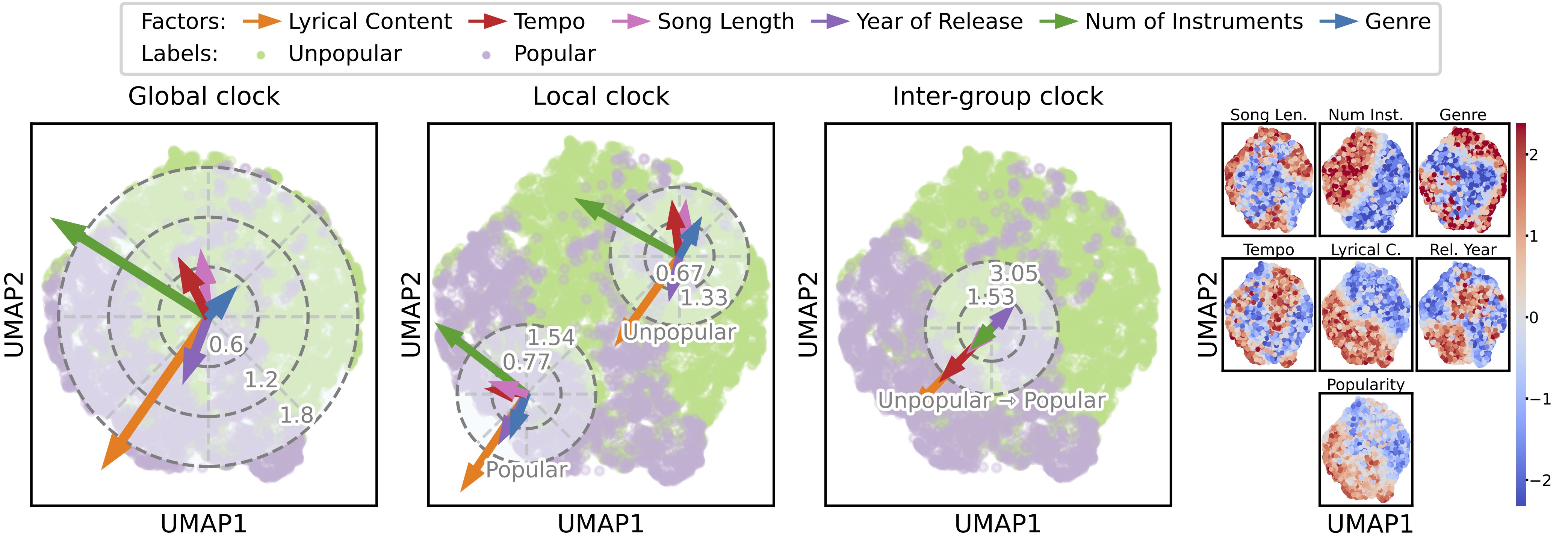}
        };
        \node at (-7.9cm,1.26cm){{a}};
        \node at (-3.83cm,1.26cm){{b}};
        \node at (0.25cm,1.26cm){{c}};
        \node at (4.3cm,1.26cm){{d}};
    \end{tikzpicture}
    \caption{To explore melody popularity, we visualize (a)~a global clock for all songs, (b)~a Local Clock for each class, and (c)~an inter-group clock between two classes. (d)~For each feature, we show a scatter to confirm the correct behaviour of clocks where  \textit{Popularity} is the original continuous feature. We transformed \textit{Popularity} into two classes for the clocks.\label{fig:melodyClocks}}
\end{figure*}

\subsection{Neftel Dataset}
\label{appendix:neftel}

We use a dataset popularly called Neftel~\cite{NEFTEL}.
The dataset includes a single-cell analysis of glioblastoma, a type of cancer that starts as a growth of cells in the brain or spinal cord. 
The dataset provides a blueprint for glioblastoma, integrating the malignant cell programs and their modulation by genetic drivers.
There are six glioblastoma state signatures for malignant cells included in the data: NPClike1, NPClike2, MESlike1, MESlike2, AClike, and OPlike. 
Given all signatures and the number of expressed genes, biologists try to detect which cells (data points) are characterized by which signature.

A usual approach to compare signatures is to plot low-dimensional data for each signature color-coding data point with their high-dimensional values, as shown in the rightmost plot in~\cref{fig:neftelClocks}.
Another method is to assign labels to each point and to visualize low-dimensional data color-coded by these labels.
Labels are created by assigning the name of a signature with maximum value.
With this method, a user can see if there are any clusters or structures in the data.

\cref{fig:neftelClocks} shows global, local, and inter-group clocks.
The Global Clock gives a compact overview of general trends in the data.
We see that the clock in~\cref{appendix:neftel}a shows how values change and that, e.g., OPClike signatures have the biggest impact in approximately $70\degree$ direction.
The same trend can be seen in scatter plots.

Two local clocks analyze embeddings at a finer granularity. We use HDBSCAN to find clusters in high-dimensional data.
There are two clusters, and, therefore, two clocks.
For the big cluster on the left, the direction of impact of the OPClike signatures slightly changes compared to the big clock. 
In the second clock (on the right), an OPClike arrow points in the direction of $150\degree$.
Properties of points of the right cluster differ from the left one, this leads to completely different directions and higher coefficients (arrow length marked on circle).
This can be interpreted in connection with the Global Clock where the effect of combining impacts of both clusters is visualized.
Although the right cluster has bigger coefficients, it is impact on the Global Clock is leveled out because of the smaller amount of points.
Such details as the impact within a single cluster can not be seen on a bigger scale and even when plotting all features.

\cref{fig:neftelClocks}c, the Inter-group Clock, shows the relationship between two HDBSCAN clusters and which features can be changed to move from the center of one cluster to the center of another.
It is visible that the most significant feature between the two clusters is NPClike2.
NPClike2 arrow points in the direction of a smaller cluster (on the right).
This means that to move from a bigger cluster to a smaller one, we need to increase NPClike2 impact.
The same trend can be seen in the rightmost plot in the~\cref{fig:neftelClocks}.
Additionally, scatter plots justify that the biggest value difference between the points is in NPClike2 signatures.

\subsection{Pima Indians Diabetes Database}
\label{appendix:pima}

The Pima Indians Diabetes Database~\cite{pimaData} originates from the National Institute of Diabetes and Digestive and Kidney Diseases. 
The goal of the dataset is to predict if a patient has diabetes based on personal data, e.g., blood pressure and skin sickness.
We use actual class labels for the Local and Inter-group Clocks.

In the Global Clock, \cref{fig:pimaClocks}a, we see that blood pressure, age, glucose, and insulin level influence diabetes among Pima females.
Interestingly, skin thickness varies among all observations, but body mass index (BMI) is insignificant.

\cref{fig:pimaClocks}b shows Local Clocks for two classes. Both clocks are driven by almost the same features but the directions differ. 
Additionally, BMI is insignificant for the both classes.

Inter-group Clock highlights that when pedigree, BMI, and glucose increase, the likelihood of having diabetes is higher.

\subsection{Melody Metrics: Decoding Song Popularity}
\label{appendix:melody}

A synthetic dataset~\cite{melody_data} from Kaggle is designed to unravel the mystique behind song popularity. 
The task is to predict song popularity (regression) based on the information about it.
We create two classes from of the original target \textit{Popularity} (see~\cref{fig:melodyClocks}d), and visualize a clocks for all songs.

From the Global Clock in~\cref{fig:melodyClocks}a, it is clear that lyrical content and a more recent year of release distinguish popular songs.
Number of instruments has the highest variability among all observations. 
Genre, song length, and tempo influence music towards lower popularity, but contributions of these features are significantly lower compared with a number of instruments.

Local Clocks, \cref{fig:melodyClocks}b, show that same features drive both classes but some variable point in different directions.

The Inter-group Clock in~\cref{fig:melodyClocks}c provides insights into what distinguishes unpopular melodies from hits. We can "move" between two classes by changing lyrical content, tempo, year of release, or number of instruments. Additionally, it appears that changes of genre are insignificant between two classes.

%% file: figures/circleTheoremTikz.tex
\setlength{\columnsep}{0.3cm}%
\begin{wrapfigure}{l}{4cm}
    \centering
    \begin{tikzpicture}[scale=1.1]
        \draw[gray, thick, ->] (-1,0) -- (2.3,0);
        \draw[gray, thick, ->] (0,-0.6) -- (0,2);

        \draw[thick] (0.8,0.6) circle (1);

        \filldraw[gray] (0,0) circle (2pt);
        \node[anchor=north east](O)at(0,0){$O$};
        \filldraw[gray](0,1.2) circle (2pt);
        \node[anchor=south east](B)at(0,1.2){$B$};
        \filldraw[gray](1.6,0) circle (2pt);
        \node[anchor=north west](D)at(1.6,0){$D$};

        \draw[blue, thick, -](0,0) -- (1.76,0.88);

        \filldraw[blue](1.76,0.88) circle(2pt);
        \node[anchor=north west, blue](C)at(1.76,0.88){$C$};

        \filldraw[gray](0.8,0.6) circle(0.5pt);

        \filldraw[red](1.6,1.2) circle(2pt);
        \node[anchor=south west, red](K)at(1.28,1.1){$K$};
        \draw[red, thick, -](0,0) -- (1.6,1.2);
            
        \draw[orange, thick, -](B.south east) -- (D.north west);
        \draw[orange, thick, -](B.south east) -- (C.north west);
        \draw[orange, thick, ]
            (0.2296,1.15825) .. controls (0.2277,1.099)..
            (0.1890,1.05823);
        
        \draw[blue, thick, ]
            (0.2296,0) .. controls (0.2277,0.0519)..
            (0.1890,0.0945);
        \node[anchor=west, blue] at(0.3277,0.119){$\theta$};

        \begin{scope}[on background layer]      
            \draw[gray](1.54888,0.91838) -- (1.51198,0.715425);
            \draw[gray](1.51198,0.715425) -- (1.72309,0.677041);
            \draw[gray](C.north west) -- (D.north west);
        \end{scope}

        \draw[green, dashed, thick](0,0) -- (C.north west);
        \node[green]at (0.7315,0.270){$d$};
         
    \end{tikzpicture}

    \caption{Single projection}
    \label{fig:circleTheorem}
\end{wrapfigure}

%% file: main.bbl
\begin{thebibliography}{10}

\bibitem{berisha2021digital}
V.~Berisha, C.~Krantsevich, P.~R. Hahn, S.~Hahn, G.~Dasarathy, P.~Turaga, and J.~Liss.
\newblock Digital medicine and the curse of dimensionality.
\newblock {\em npj Digital Medicine}, 4(1):153, 2021. doi: {{%
10\hspace{.1pt}\discretionary{.}{%
}{.}\hspace{.4pt}1038\discretionary{/}{%
}{/}s41746\discretionary{%
}{-}{-}021\discretionary{%
}{-}{-}00521\discretionary{%
}{-}{-}5}}


\bibitem{tviSNE}
A.~Chatzimparmpas, R.~M. Martins, and A.~Kerren.
\newblock {t-viSNE: Interactive Assessment and Interpretation of t-SNE Projections}.
\newblock {\em {IEEE Transactions on Visualization and Computer Graphics}}, 26(8):2696--2714, 2020. doi: {{%
10\hspace{.1pt}\discretionary{.}{%
}{.}\hspace{.4pt}1109\discretionary{/}{%
}{/}TVCG\hspace{.1pt}\discretionary{.}{%
}{.}\hspace{.4pt}2020\hspace{.1pt}\discretionary{.}{%
}{.}\hspace{.4pt}2986996}}


\bibitem{Cheng2013HW}
M.-Y. Cheng and H.~tieng Wu.
\newblock Local linear regression on manifolds and its geometric interpretation.
\newblock {\em Journal of the American Statistical Association}, 2013. doi: {{%
10\hspace{.1pt}\discretionary{.}{%
}{.}\hspace{.4pt}1080\discretionary{/}{%
}{/}01621459\hspace{.1pt}\discretionary{.}{%
}{.}\hspace{.4pt}2013\hspace{.1pt}\discretionary{.}{%
}{.}\hspace{.4pt}827984}}


\bibitem{Cheng2016TheDC}
S.~Cheng and K.~Mueller.
\newblock The data context map: Fusing data and attributes into a unified display.
\newblock {\em IEEE Transactions on Visualization and Computer Graphics}, 22:121--130, 2016. doi: {{%
10\hspace{.1pt}\discretionary{.}{%
}{.}\hspace{.4pt}1109\discretionary{/}{%
}{/}TVCG\hspace{.1pt}\discretionary{.}{%
}{.}\hspace{.4pt}2015\hspace{.1pt}\discretionary{.}{%
}{.}\hspace{.4pt}2467552}}


\bibitem{support2}
A.~F. e.~a. Connors.
\newblock {A Controlled Trial to Improve Care for Seriously III Hospitalized Patients: The Study to Understand Prognoses and Preferences for Outcomes and Risks of Treatments (SUPPORT)}.
\newblock {\em JAMA}, pp. 1591--1598, 1995. doi: {{%
10\hspace{.1pt}\discretionary{.}{%
}{.}\hspace{.4pt}1001\discretionary{/}{%
}{/}jama\hspace{.1pt}\discretionary{.}{%
}{.}\hspace{.4pt}1995\hspace{.1pt}\discretionary{.}{%
}{.}\hspace{.4pt}03530200027032}}


\bibitem{Eckelt2023Visual}
K.~Eckelt, A.~Hinterreiter, P.~Adelberger, C.~Walchshofer, V.~Dhanoa, C.~Humer, M.~Heckmann, C.~Steinparz, and M.~Streit.
\newblock Visual exploration of relationships and structure in low-dimensional embeddings.
\newblock {\em IEEE Transactions on Visualization and Computer Graphics}, 29(7):3312--3326, 2023. doi: {{%
10\hspace{.1pt}\discretionary{.}{%
}{.}\hspace{.4pt}1109\discretionary{/}{%
}{/}TVCG\hspace{.1pt}\discretionary{.}{%
}{.}\hspace{.4pt}2022\hspace{.1pt}\discretionary{.}{%
}{.}\hspace{.4pt}3156760}}


\bibitem{Faust2019DimReader}
R.~Faust, D.~Glickenstein, and C.~Scheidegger.
\newblock Dimreader: Axis lines that explain non-linear projections.
\newblock {\em IEEE Transactions on Visualization and Computer Graphics}, pp. 481--490, 2019. doi: {{%
10\hspace{.1pt}\discretionary{.}{%
}{.}\hspace{.4pt}1109\discretionary{/}{%
}{/}TVCG\hspace{.1pt}\discretionary{.}{%
}{.}\hspace{.4pt}2018\hspace{.1pt}\discretionary{.}{%
}{.}\hspace{.4pt}2865194}}


\bibitem{IrisData}
R.~Fisher.
\newblock Iris flower dataset, 1936.

\bibitem{Fujiwara2020Supporting}
T.~Fujiwara, O.~Kwon, and K.~Ma.
\newblock Supporting analysis of dimensionality reduction results with contrastive learning.
\newblock {\em IEEE Transactions on Visualization and Computer Graphics}, 26(01):45--55, 2020. doi: {{%
10\hspace{.1pt}\discretionary{.}{%
}{.}\hspace{.4pt}1109\discretionary{/}{%
}{/}TVCG\hspace{.1pt}\discretionary{.}{%
}{.}\hspace{.4pt}2019\hspace{.1pt}\discretionary{.}{%
}{.}\hspace{.4pt}2934251}}


\bibitem{biplot}
K.~R. Gabriel.
\newblock The biplot graphic display of matrices with application to principal component analysis.
\newblock {\em Biometrika}, 3:453--467, 1971. doi: {{%
10\hspace{.1pt}\discretionary{.}{%
}{.}\hspace{.4pt}2307\discretionary{/}{%
}{/}2334381}}


\bibitem{Gorin2022RNA}
G.~Gorin, M.~Fang, T.~Chari, and L.~Pachter.
\newblock Rna velocity unraveled.
\newblock {\em bioRxiv}, 2022. doi: {{%
10\hspace{.1pt}\discretionary{.}{%
}{.}\hspace{.4pt}1101\discretionary{/}{%
}{/}2022\hspace{.1pt}\discretionary{.}{%
}{.}\hspace{.4pt}02\hspace{.1pt}\discretionary{.}{%
}{.}\hspace{.4pt}12\hspace{.1pt}\discretionary{.}{%
}{.}\hspace{.4pt}480214}}


\bibitem{indyk1998approximate}
P.~Indyk and R.~Motwani.
\newblock {Approximate Nearest Neighbors: Towards Removing the Curse of Dimensionality}.
\newblock In {\em ACM Symposium on Theory of Computing}, p. 604–613, 1998. doi: {{%
10\hspace{.1pt}\discretionary{.}{%
}{.}\hspace{.4pt}1145\discretionary{/}{%
}{/}276698\hspace{.1pt}\discretionary{.}{%
}{.}\hspace{.4pt}276876}}


\bibitem{jaradat2021tutorial}
Y.~Jaradat, M.~Masoud, I.~Jannoud, et~al.
\newblock {A Tutorial on Singular Value Decomposition with Applications on Image Compression and Dimensionality Reduction}.
\newblock In {\em International Conference on Information Technology}, 2021. doi: {{%
10\hspace{.1pt}\discretionary{.}{%
}{.}\hspace{.4pt}1109\discretionary{/}{%
}{/}ICIT52682\hspace{.1pt}\discretionary{.}{%
}{.}\hspace{.4pt}2021\hspace{.1pt}\discretionary{.}{%
}{.}\hspace{.4pt}9491732}}


\bibitem{Joia2015Uncovering}
P.~Joia, F.~Petronetto, and L.~G. Nonato.
\newblock {Uncovering Representative Groups in Multidimensional Projections}.
\newblock {\em Computer Graphics Forum}, 2015. doi: {{%
10\hspace{.1pt}\discretionary{.}{%
}{.}\hspace{.4pt}1111\discretionary{/}{%
}{/}cgf\hspace{.1pt}\discretionary{.}{%
}{.}\hspace{.4pt}12640}}


\bibitem{melody_data}
K.~Karuarathna.
\newblock Melody metrics: Decoding song popularity, 2023. doi: {{%
10\hspace{.1pt}\discretionary{.}{%
}{.}\hspace{.4pt}34740\discretionary{/}{%
}{/}KAGGLE\discretionary{/}{%
}{/}DS\discretionary{/}{%
}{/}3914464}}


\bibitem{KL}
S.~Kullback and R.~A. Leibler.
\newblock On information and sufficiency.
\newblock {\em The Annals of Mathematical Statistics}, pp. 79--86, 1951. doi: {{%
10\hspace{.1pt}\discretionary{.}{%
}{.}\hspace{.4pt}1214\discretionary{/}{%
}{/}aoms\discretionary{/}{%
}{/}1177729694}}


\bibitem{Virgilio22}
C.~Li, M.~Virgilio, K.~Collins, and J.~Welch.
\newblock Multi-omic single-cell velocity models epigenome–transcriptome interactions and improves cell fate prediction.
\newblock {\em Nature Biotechnology}, 41:1--12, 10 2022. doi: {{%
10\hspace{.1pt}\discretionary{.}{%
}{.}\hspace{.4pt}1038\discretionary{/}{%
}{/}s41587\discretionary{%
}{-}{-}022\discretionary{%
}{-}{-}01476\discretionary{%
}{-}{-}y}}


\bibitem{Zhangyu23QY}
Z.~Li and Y.~Qiu.
\newblock Feature selection based on improved principal component analysis.
\newblock In {\em Conference on Algorithms, Computing and Machine Learning}, p. 188–192, 2023. doi: {{%
10\hspace{.1pt}\discretionary{.}{%
}{.}\hspace{.4pt}1145\discretionary{/}{%
}{/}3590003\hspace{.1pt}\discretionary{.}{%
}{.}\hspace{.4pt}3590036}}


\bibitem{Linderman2017ClusteringWT}
G.~C. Linderman and S.~Steinerberger.
\newblock {Clustering with t-SNE, provably}.
\newblock {\em SIAM Journal on Mathematics of Data Science}, 1 2:313--332, 2017. doi: {{%
10\hspace{.1pt}\discretionary{.}{%
}{.}\hspace{.4pt}1137\discretionary{/}{%
}{/}18M1216134}}


\bibitem{Liu2017Visualizing}
S.~Liu, D.~Maljovec, B.~Wang, P.-T. Bremer, and V.~Pascucci.
\newblock Visualizing high-dimensional data: Advances in the past decade.
\newblock {\em IEEE Transactions on Visualization and Computer Graphics}, pp. 1249--1268, 2017. doi: {{%
10\hspace{.1pt}\discretionary{.}{%
}{.}\hspace{.4pt}1109\discretionary{/}{%
}{/}TVCG\hspace{.1pt}\discretionary{.}{%
}{.}\hspace{.4pt}2016\hspace{.1pt}\discretionary{.}{%
}{.}\hspace{.4pt}2640960}}


\bibitem{ManduchiMarcinkevics2022}
L.~Manduchi, R.~Marcinkevi{\v{c}}s, M.~C. Massi, T.~Weikert, A.~Sauter, V.~Gotta, T.~M{\"u}ller, F.~Vasella, M.~C. Neidert, M.~Pfister, B.~Stieltjes, and J.~E. Vogt.
\newblock A deep variational approach to clustering survival data.
\newblock In {\em International Conference on Learning Representations}, 2022.

\bibitem{MarcilioJr2021_ClusterShapley}
W.~E. Marcílio-Jr and D.~M. Eler.
\newblock Explaining dimensionality reduction results using shapley values.
\newblock {\em Expert Systems with Applications}, 178:115020, 2021. doi: {{%
10\hspace{.1pt}\discretionary{.}{%
}{.}\hspace{.4pt}1016\discretionary{/}{%
}{/}j\hspace{.1pt}\discretionary{.}{%
}{.}\hspace{.4pt}eswa\hspace{.1pt}\discretionary{.}{%
}{.}\hspace{.4pt}2021\hspace{.1pt}\discretionary{.}{%
}{.}\hspace{.4pt}115020}}


\bibitem{UMAP}
L.~{McInnes}, J.~{Healy}, and J.~{Melville}.
\newblock {UMAP: Uniform Manifold Approximation and Projection for Dimension Reduction}.
\newblock {\em ArXiv e-prints}, 2018.

\bibitem{PHATE}
K.~R. Moon, D.~van Dijk, S.~G. Zheng~Wang, et~al.
\newblock Visualizing structure and transitions in high-dimensional biological data.
\newblock {\em Nature Biotechnology}, 37:1482--1492, 2019. doi: {{%
10\hspace{.1pt}\discretionary{.}{%
}{.}\hspace{.4pt}1038\discretionary{/}{%
}{/}s41587\discretionary{%
}{-}{-}019\discretionary{%
}{-}{-}0336\discretionary{%
}{-}{-}3}}


\bibitem{NEFTEL}
C.~Neftel et~al.
\newblock An integrative model of cellular states, plasticity, and genetics for glioblastoma.
\newblock {\em Cell}, 178(4):835--849.e21, 2019. doi: {{%
10\hspace{.1pt}\discretionary{.}{%
}{.}\hspace{.4pt}1016\discretionary{/}{%
}{/}j\hspace{.1pt}\discretionary{.}{%
}{.}\hspace{.4pt}cell\hspace{.1pt}\discretionary{.}{%
}{.}\hspace{.4pt}2019\hspace{.1pt}\discretionary{.}{%
}{.}\hspace{.4pt}06\hspace{.1pt}\discretionary{.}{%
}{.}\hspace{.4pt}024}}


\bibitem{Prummer23}
M.~Prummer, A.~Bertolini, L.~Bosshard, F.~Barkmann, J.~Yates, V.~Boeva, T.~T.~P. Consortium, D.~Stekhoven, and F.~Singer.
\newblock {scROSHI: robust supervised hierarchical identification of single cells}.
\newblock {\em NAR Genomics and Bioinformatics}, 2023. doi: {{%
10\hspace{.1pt}\discretionary{.}{%
}{.}\hspace{.4pt}1093\discretionary{/}{%
}{/}nargab\discretionary{/}{%
}{/}lqad058}}


\bibitem{SainburgML2020}
T.~Sainburg, L.~McInnes, and T.~Q. Gentner.
\newblock {Parametric UMAP: learning embeddings with deep neural networks for representation and semi-supervised learning}.
\newblock {\em ArXiv e-prints}, 2021.

\bibitem{pimaData}
J.~Smith, J.~Everhart, W.~Dickson, W.~Knowler, and R.~Johannes.
\newblock {Using the ADAP Learning Algorithm to Forcast the Onset of Diabetes Mellitus}.
\newblock {\em Annual Symposium on Computer Applications in Medical Care}, 10, 11 1988.

\bibitem{Sohns2022Attribute}
J.-T. Sohns, M.~Schmitt, F.~Jirasek, H.~Hasse, and H.~Leitte.
\newblock Attribute-based explanation of non-linear embeddings of high-dimensional data.
\newblock {\em IEEE Transactions on Visualization and Computer Graphics}, pp. 540--550, 2022. doi: {{%
10\hspace{.1pt}\discretionary{.}{%
}{.}\hspace{.4pt}1109\discretionary{/}{%
}{/}TVCG\hspace{.1pt}\discretionary{.}{%
}{.}\hspace{.4pt}2021\hspace{.1pt}\discretionary{.}{%
}{.}\hspace{.4pt}3114870}}


\bibitem{Stahnke2016Probing}
J.~Stahnke, M.~Dörk, B.~Müller, and A.~Thom.
\newblock Probing projections: Interaction techniques for interpreting arrangements and errors of dimensionality reductions.
\newblock {\em IEEE Transactions on Visualization and Computer Graphics}, 22(1):629--638, 2016. doi: {{%
10\hspace{.1pt}\discretionary{.}{%
}{.}\hspace{.4pt}1109\discretionary{/}{%
}{/}TVCG\hspace{.1pt}\discretionary{.}{%
}{.}\hspace{.4pt}2015\hspace{.1pt}\discretionary{.}{%
}{.}\hspace{.4pt}2467717}}


\bibitem{Turkay2012Representative}
C.~Turkay, A.~Lundervold, A.~J. Lundervold, and H.~Hauser.
\newblock Representative factor generation for the interactive visual analysis of high-dimensional data.
\newblock {\em IEEE Transactions on Visualization and Computer Graphics}, 18(12):2621--2630, 2012. doi: {{%
10\hspace{.1pt}\discretionary{.}{%
}{.}\hspace{.4pt}1109\discretionary{/}{%
}{/}TVCG\hspace{.1pt}\discretionary{.}{%
}{.}\hspace{.4pt}2012\hspace{.1pt}\discretionary{.}{%
}{.}\hspace{.4pt}256}}


\bibitem{tSNE}
L.~van~der Maaten and G.~Hinton.
\newblock Visualizing data using t-sne.
\newblock {\em Journal of Machine Learning Research}, 9(86):2579--2605, 2008.

\bibitem{wattenberg2016how}
M.~Wattenberg, F.~Viégas, and I.~Johnson.
\newblock How to use t-sne effectively.
\newblock {\em Distill}, 2016. doi: {{%
10\hspace{.1pt}\discretionary{.}{%
}{.}\hspace{.4pt}23915\discretionary{/}{%
}{/}distill\hspace{.1pt}\discretionary{.}{%
}{.}\hspace{.4pt}00002}}


\bibitem{Yates22VB}
J.~Yates and V.~Boeva.
\newblock {Deciphering the etiology and role in oncogenic transformation of the CpG island methylator phenotype: a pan-cancer analysis}.
\newblock {\em Briefings in Bioinformatics}, 2022. doi: {{%
10\hspace{.1pt}\discretionary{.}{%
}{.}\hspace{.4pt}1093\discretionary{/}{%
}{/}bib\discretionary{/}{%
}{/}bbab610}}


\bibitem{PCACompression}
J.~Ye, R.~Janardan, and Q.~Li.
\newblock Gpca: An efficient dimension reduction scheme for image compression and retrieval.
\newblock In {\em Proceedings of the ACM SIGKDD International Conference on Knowledge Discovery and Data Mining}, pp. 354--363, 2004. doi: {{%
10\hspace{.1pt}\discretionary{.}{%
}{.}\hspace{.4pt}1145\discretionary{/}{%
}{/}1014052\hspace{.1pt}\discretionary{.}{%
}{.}\hspace{.4pt}1014092}}


\bibitem{Hongfang19ZY}
H.~Zhou, Y.~Zhang, Y.~Zhang, and H.~Liu.
\newblock Feature selection based on conditional mutual information: minimum conditional relevance and minimum conditional redundancy.
\newblock {\em Applied Intelligence}, 2019. doi: {{%
10\hspace{.1pt}\discretionary{.}{%
}{.}\hspace{.4pt}1007\discretionary{/}{%
}{/}s10489\discretionary{%
}{-}{-}018\discretionary{%
}{-}{-}1305\discretionary{%
}{-}{-}0}}


\end{thebibliography}
